\documentclass[final,3p,times]{elsarticle}

\usepackage{amssymb}
\usepackage{amsmath}
\usepackage{array}
\usepackage{longtable}

\usepackage{hyperref}
\usepackage{multirow}
\journal{Expert Systems With Applications}
 
\usepackage[dvipsnames,table]{xcolor}
\usepackage{soul,etoolbox}
\setstcolor{blue}
\usepackage{relsize}
\usepackage{multirow}

\colorlet{punct}{red!60!black}
\definecolor{background}{HTML}{EEEEEE}
\definecolor{c}{gray}{0.9}
\definecolor{delim2}{RGB}{20,105,176}
\definecolor{delim}{RGB}{148, 49, 38}

\colorlet{numb}{magenta!60!black}

\usepackage{listings}%
\lstdefinelanguage{python}{
basicstyle=\small\ttfamily,
    tabsize=2,
    showstringspaces=false,
    breaklines=true,
    frame=lines,
    backgroundcolor=\color{gray!10},
    linewidth=\columnwidth,
    escapechar={*}, 
}

\usepackage{tikz}

\usepackage{varwidth}

\newcommand{\expressions}[1]{%
  \tikz[baseline={(a.base)}]%
    \node[draw=blue!40!white, line width=0.7pt, rounded corners=0.8ex, fill=blue!10!white, inner sep=2pt, text=black] (a)%
    {\begin{varwidth}{\linewidth}\raggedright #1\end{varwidth}};%
}

\newcommand{\covid}[1]{%
  \tikz[baseline={(a.base)}]%
    \node[draw=red!75!black, line width=0.7pt, rounded corners=0.8ex, fill=red!10!white, inner sep=1.5pt, text=black] (a)%
    {\begin{varwidth}{\linewidth}\raggedright #1\end{varwidth}};%
}

\newcommand{\public}[1]{%
  \tikz[baseline={(a.base)}]%
    \node[draw=orange!75!black, line width=0.7pt, rounded corners=0.8ex, fill=orange!10!white, inner sep=2pt, text=black] (a)%
    {\begin{varwidth}{\linewidth}\raggedright #1\end{varwidth}};%
}

\newcommand{\media}[1]{%
  \tikz[baseline={(a.base)}]%
    \node[draw=pink, line width=0.7pt, rounded corners=0.8ex, fill=pink!25!white, inner sep=2pt, text=black] (a)%
    {\begin{varwidth}{\linewidth}\raggedright #1\end{varwidth}};%
}

\newcommand{\social}[1]{%
  \tikz[baseline={(a.base)}]%
    \node[draw=green!75!black, line width=0.7pt, rounded corners=0.8ex, fill=green!10!white, inner sep=2pt, text=black] (a)%
    {\begin{varwidth}{\linewidth}\raggedright #1\end{varwidth}};%
}

\newcommand{\inter}[1]{%
  \tikz[baseline={(a.base)}]%
    \node[draw=brown!75!black, line width=0.7pt, rounded corners=0.8ex, fill=brown!10!white, inner sep=2pt, text=black] (a)%
    {\begin{varwidth}{\linewidth}\raggedright #1\end{varwidth}};%
}

\newcommand{\politic}[1]{%
  \tikz[baseline={(a.base)}]%
    \node[draw=black!75!black, line width=0.7pt, rounded corners=0.8ex, fill=black!70!white, inner sep=2pt, text=white] (a)%
    {\begin{varwidth}{\linewidth}\raggedright #1\end{varwidth}};%
}

\newcommand{\spain}[1]{%
  \tikz[baseline={(a.base)}]%
    \node[draw=violet!75!black, line width=0.7pt, rounded corners=0.8ex, fill=violet!11!white, inner sep=2pt, text=black] (a)%
    {\begin{varwidth}{\linewidth}\raggedright #1\end{varwidth}};%
}

\newcommand{\transport}[1]{%
  \tikz[baseline={(a.base)}]%
    \node[draw=gray!75!black, line width=0.7pt, rounded corners=0.8ex, fill=gray!11!white, inner sep=2pt, text=black] (a)%
    {\begin{varwidth}{\linewidth}\raggedright #1\end{varwidth}};%
}

\usepackage[bb=boondox,bbscaled=.95,cal=boondoxo]{mathalfa}

\usepackage{amssymb}
\usepackage{pifont}
\definecolor{DarkGreen}{rgb}{0.0, 0.5, 0.0}
\newcommand{\greencheck}{{\color{DarkGreen}\checkmark}}
\newcommand{\xmark}{\color{red}\ding{55}}%

\begin{document}

\begin{frontmatter}




\title{Exploring the topics, sentiments and hate speech\\in the Spanish information environment}


\author[umu]{Alejandro~Buitrago~L\'opez\corref{cor1}}
\ead{alejandro.buitragol@um.es}

\author[umu]{Javier~Pastor-Galindo}
\ead{javierpg@um.es}

\author[umu]{Jos\'e~A.~Ruip\'erez-Valiente}
\ead{jruiperez@um.es}

\cortext[cor1]{Corresponding author.}

\affiliation[umu]{organization={Dept. of Information and Communications Engineering, University of Murcia},
            addressline={C. Campus de Espinardo}, 
            city={Murcia},
            postcode={30100}, 
            country={Spain}}

\begin{abstract}

In the digital era, the internet and social media have transformed communication but have also facilitated the spread of hate speech and disinformation, leading to radicalization, polarization, and toxicity. This is especially concerning for media outlets due to their significant role in shaping public discourse. This study examines the topics, sentiments, and hate prevalence in $337,807$ response messages (website comments and tweets) to news from five Spanish media outlets (La Vanguardia, ABC, El Pa\'is, El Mundo, and 20 Minutos) in January 2021. These public reactions were originally labeled as distinct types of hate by experts following an original procedure, and they are now classified into three sentiment values (negative, neutral, or positive) and main topics. The BERTopic unsupervised framework was used to extract $81$ topics, manually named with the help of Large Language Models (LLMs) and grouped into nine primary categories.

Results show social issues ($22.22\%$), expressions and slang ($20.35\%$), and political issues ($11.80\%$) as the most discussed. Content is mainly negative ($62.7\%$) and neutral ($28.57\%$), with low positivity ($8.73\%$). Toxic narratives relate to conversation expressions, gender, feminism, and COVID-19. Despite low levels of hate speech ($3.98\%$), the study confirms high toxicity in online responses to social and political topics.

\end{abstract}



\begin{keyword}
Spanish mass digital media \sep Information environment \sep Polarization \sep Radicalization \sep Toxicity \sep Public reactions


\end{keyword}

\end{frontmatter}


\section{Introduction}

The rise of the internet has fundamentally transformed how individuals consume news and communicate, especially through social media. In societies valuing freedom of expression, individuals now frequently express and share their opinions, integrating this practice as a natural part of their routines. Unfortunately, this new social and informational landscape has favored an unprecedented amplification of cyber threats such as hate speech and disinformation, posing significant risks to democratic systems \cite{oficinaCT}. This situation has intensified and drawn substantial attention from the research community, governmental bodies, and the general public, particularly following extensive disinformation campaigns associated with recent events, including the COVID-19 pandemic \cite{Kim2021}, the Russia-Ukraine war \cite{Pierri2022PropagandaAM}, and the Israel-Palestine conflict \cite{apnewsMisinformationAbout}.

\begin{figure}[ht!]
	\centering{
		\includegraphics[width=0.7\textwidth]{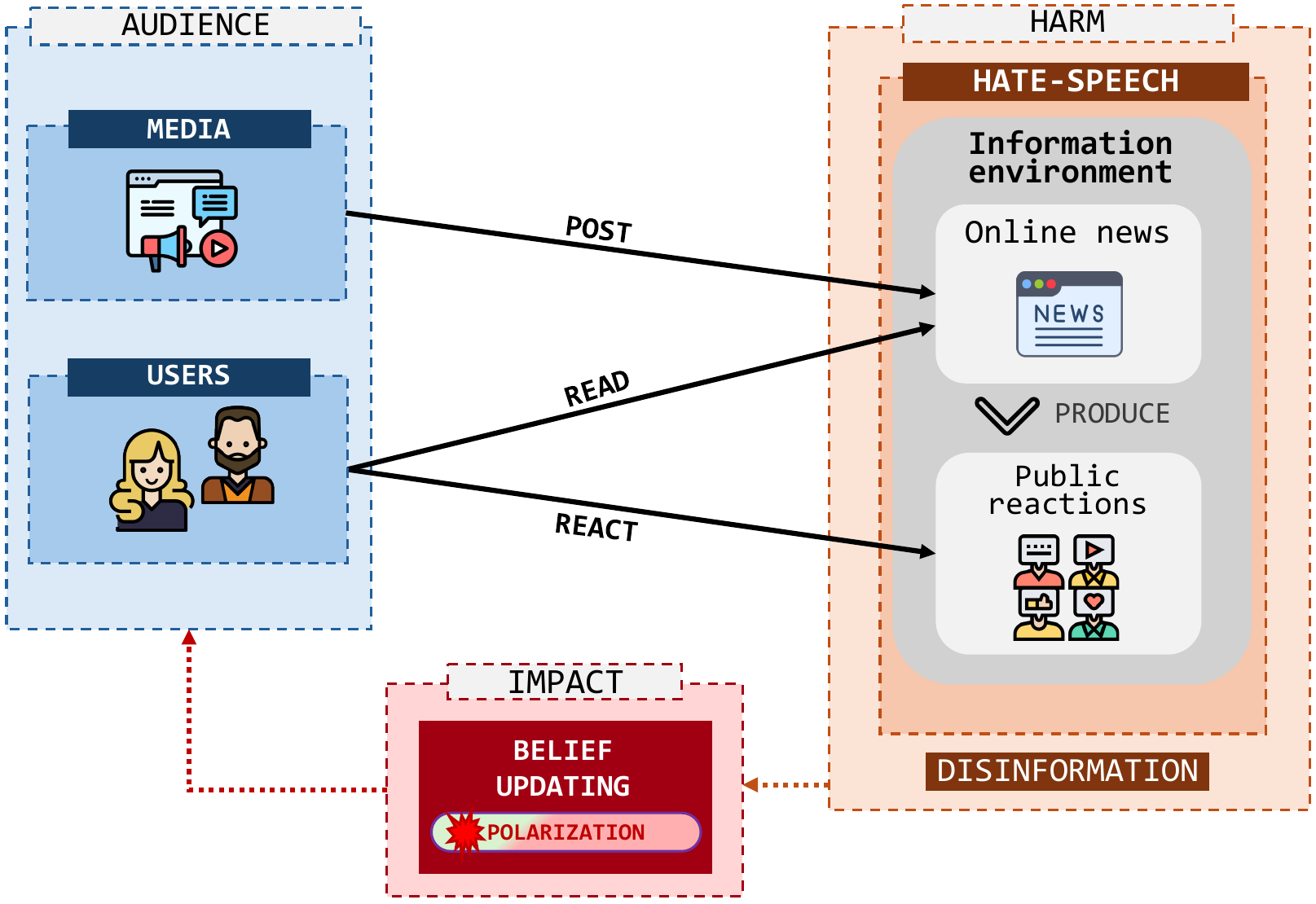}
	\caption{Threat model with media and audience (in blue), harm (in orange) and impact (in red).}\label{fig:threat-model}
 }
\end{figure}

Consequently, a structured model encapsulating the key actors, dynamics, and resulting societal impacts is proposed to understand and contextualize the environment being worked on. Figure \ref{fig:threat-model} illustrates our threat model with three main components. In blue, the media and audience as actors in the model, providing the information environment with online news and social network posts that people can read, react to, and comment on. In orange, the content is considered potentially harmful due to intrinsic hateful narratives of today's ecosystem (particularly, public reactions that will be the focus of this research work). In red, the online situation leads to polarization, extremism, and heightened tension, creating a vulnerable environment for society \cite{OSMUNDSEN_BOR_VAHLSTRUP_BECHMANN_PETERSEN_2021, Cinelli123, 9451574}. In fact, this agitated context serves as a vector for disinformation to become more effective~\cite{Kim2021}.

In this scenario, the media plays a pivotal role in delivering news that enhances the information environment~\cite{PASTORGALINDO2022161}. As online news has been characterized in the literature through several efforts, it is worth studying audience reactions to these news to understand better the publication reactions to reputable sources in terms of polarization, radicalization, and extremism, especially concerning hate speech within social networks in the case of this study.

The European Union Agency for Cybersecurity (ENISA) classifies hate speech as one of the eight major categories of cybersecurity threats in their latest report \cite{enisa2023}. One of the primary challenges in combating these threats is the sheer volume of online content~\cite{PASTORGALINDO202312}. Every day, millions of social media posts, news articles, and various other types of content are produced \cite{internet_minute_2023}. This overwhelming amount of information makes it nearly impossible for humans to detect and eliminate disinformation and hate speech manually~\cite{10492674}. In this sense, topic modeling is a machine-learning technique that enables the analysis and categorization of large textual datasets into common themes or topics \cite{Salminen2018AnatomyOO}. Additionally, since these threats often carry negative sentiments, sentiment analysis helps detect and mitigate their spread \cite{fingerprints-of-misinformation}. 

Beyond our preliminary work~\cite{JNIC24}, this study, in collaboration with the Hatemedia project~\cite{hatemedia_website}, used a dataset with $348,123$ reactions to online news from five of the most reputable Spanish mass digital media outlets (\textit{La Vanguardia}, \textit{ABC}, \textit{El Pa'is}, \textit{El Mundo}, and \textit{20 Minutos}), as identified by \textit{SCImago Media} in 2024 \cite{scimagomedia}. For simplicity, these will be referred to as Spanish media outlets throughout the rest of this paper. The data was collected from two different platforms, Twitter (currently $\mathbb{X}$) and the original web publications. This content was gathered during January 2021 and manually labeled into various types of hate between February and April 2022.

To contextualize the data from January 2021, it is essential to consider the significant events leading up to it. Following the 2020 presidential election, where Joe Biden was declared the winner, Trump and his allies attempted to contest the results, culminating in a violent insurrection at the Capitol on January 6, 2021 \cite{duignan2024january}. Concurrently, the COVID-19 pandemic persisted, marked by vaccine rollouts and debates over their distribution, efficacy, and misinformation \cite{Kim2021}. In Spain, political tensions were high due to debates over the government's handling of the pandemic and the contentious Catalan independence movement. The country also faced economic challenges due to the pandemic, particularly on tourism and unemployment. Finally, Spain experienced significant climatological events, most notably the `Storm Filomena,' which caused significant disruptions and damage \cite{nhess-2021-396}.

Under these conditions, this study extends the previous analysis conducted in the context of the Hatemedia project. It seeks to analyze public reactions in the communication ecosystem of that period, with the following objectives:

\begin{enumerate}
    \item[O1] \textbf{Identify the topics within public reactions to Spanish media content}: For understanding the main themes and issues that dominate public discourse in the Spanish information environment.
    \item[O2] \textbf{Identify the expressed sentiment (positive, negative, or neutral) in public reactions}: For understanding the emotional tone surrounding different topics in the public discourse and how different topics affect public emotions.
    \item[O3] \textbf{Analyze the prevalence of hate in public reactions}: For identifying and addressing toxic and harmful discourse that can raise social tensions.
    \item[O4] \textbf{Characterize the five Spanish media outlets}: For understanding the topics, sentiments, and hate generated by the news of the five Spanish media and how they influence public reactions.
\end{enumerate}

To the best of our knowledge, this study is the first to characterize the topics and sentiments of public reactions to professional Spanish media content, which was manually labeled according to different types of hate. The analysis includes examining topics, sentiments, and hate speech in the dataset dated January 2021 and characterizing five prominent Spanish media outlets. Identifying topics helps discern the interests and issues engaging the public discourse. Moreover, exploring the sentiment of public reactions helps assess the emotional impact of media content on the audience on different topics. Furthermore, analyzing the prevalence of hate speech helps in identifying and addressing toxic and harmful narratives. Lastly, characterizing the media outlets provides insights into how different media sources influence public opinion and discourse, understanding media bias and the role of media in shaping public reactions.

The remainder of the paper is organized as follows. In Section \ref{sota}, related works are presented. Section \ref{met} describes the methodology followed in the study. Section \ref{results} presents the results. Finally, we end the paper with conclusions and future works in Section \ref{conclusion}.

\section{State of the art}\label{sota}

The study of public reactions has become essential as individuals have unprecedented access to news content and the ability to interact with it in various ways. By way of example, in \cite{GARCIA2021107057}, the authors analyzed public reactions on Twitter during the COVID-19 pandemic in 2020 from Brazil and the USA. They focused on sentiment analysis and topic identification, using GSDMM, the CrystalFeel algorithm for English sentiment analysis, and the Multilingual Universal Sentence Encoder with logistic regression for Portuguese. They analyzed 3,332,565 tweets in English and 3,155,277 in Portuguese to provide insights into public discourse, sentiment trends, and strategic public health communication. The study identified ten main themes, with negative emotions dominating almost all. Topics ranged from case reports and statistics to politics, treatment, education, and culture. The findings suggested that most themes were similar in English and Portuguese, with negative messages prevailing.

Furthermore, in \cite{Sattar2021}, the authors employed natural language processing and sentiment analysis techniques in 1.2 million 2021 tweets to analyze people's attitudes toward the COVID-19 vaccines in the USA. In particular, for pre-processing tweets before sentiment analysis, `NLTK' was employed \cite{bird2009natural}. On the other hand, the study's authors used both `VADER' \cite{Hutto_Gilbert_2014} and `TextBlob' \cite{Loria2020} for sentiment analysis. The study concluded that people had positive sentiments towards taking COVID-19 vaccines instead of some adverse effects of some of the vaccines.

Furthermore, in \cite{WanParis2021}, the authors introduced `VizieFeel,' a system that monitors and analyzes public emotional reactions on Twitter for specific topics or events. The system utilizes emotion analysis and content analysis with $133K$ collected tweets based on queries designed to capture data related to specific topics or events. It employs a large vocabulary of emotional words mapped to an emotion hierarchy, allowing for the analysis of incoming posts and visualization of the emotional pulse of a geographical region. This system was notably used to study public reactions to the Sydney siege tragedy in 2014, revealing an unexpected increase in joy alongside sadness. The content analysis highlighted messages of hope, praise for the police, prayers, and unity expressions from a broad community spectrum.

Regarding public reactions to media outlets' content, in \cite{Fiesler2018}, the authors examined the public reactions to data sharing and privacy controversies by analyzing comments on news articles about `WhatsApp' sharing data with `Facebook' and `unroll.me' selling anonymized data to Uber. The results obtained in this study indicate that public reactions to data privacy controversies are complex and multifaceted. Users have nuanced attitudes toward privacy, with divided opinions on who should protect data: some believe that users should bear this responsibility. In contrast, others think that companies should do so. Moreover, public outrage relates more to pre-existing notions of online privacy than to the specific nature of the breach.

Additionally, the authors in \cite{MAQSA} presented the platform called ``MAQSA'' that provides an interactive topic-centric dashboard summarizing news articles and social activity, including comments and tweets, around them. This platform also includes sentiment analysis on news article comments to group them by user opinion. They employed LDA, TF-IDF, and Open-Calais to extract topics and entities from article titles and contents. The study aimed to provide a social analytics system for newsrooms, assisting editors and publishers in understanding audience sentiment and user engagement across various topics. Additionally, it enabled news consumers to explore public reactions to articles and refine their exploration based on related entities, topics, articles, and tweets.

Moreover, in \cite{Goodwin2018848}, the authors analyzed public responses to online news about a water reuse proposal in London to understand perceptions, concerns, and attitudes. Using framing analysis, they examine whether media frames resonate with popular knowledge and influence public reactions. Techniques like pattern-matching compare themes and sentiment strength between articles and comments, with quantitative outputs such as the proportion of comments per theme aiding interpretation. The findings suggest that preexisting public perceptions and attitudes impact reactions to water reuse schemes more than media framing, highlighting the importance of effective communication to improve public support.

Furthermore, in \cite{CHOI201750}, the authors discussed the analysis of public reactions and sentiments during infectious disease outbreaks, particularly focusing on the Middle East respiratory syndrome coronavirus (MERS) outbreak in Korea in 2015. Sentiment analysis, text mining, and machine learning were employed to understand the dynamics of information flow between disease outbreaks, media coverage, and public reactions. The study highlights the significant impact of media coverage on shaping public reactions and emphasizes the need for better management of public sentiments during health crises to prevent overreactions.

In this vein, the authors in \cite{CHEN20151780} examined media and public reactions and sentiments to the online news reports of suspected vaccine adverse events and relevant policy changes in the interactive media environment to better understand how these reactions develop and how they can influence public perception and confidence in vaccination. Public reactions are analyzed through Internet monitoring of online media coverage, `Sina Weibo' postings, and `Baidu search' engine indexes. In addition, sentiment is analyzed by human coding of original posts as negative, neutral, or positive. The study concludes that government responses and policy changes influenced public sentiment, with negative publications dominating and underscoring the importance of timely and transparent communication from authorities to address negative sentiment and prevent misinformation during vaccine crises in the digital age.  

Within the Spanish context, in \cite{Miranda2023}, public reactions are explored through sentiment analysis of Spanish pandemic tweets using a fine-tuned BERT architecture, specifically emphasizing distinguishing between positive, negative, and neutral sentiments. This study reveals that the number of positive tweets was significantly lower than the other categories. The predominant emotions identified in the Spanish tweets about COVID-19 were anger, followed by joy. Sadness and fear also appeared as relevant sentiments. Finally, in \cite{9226407}, the authors examined Spanish tweets during the 2019 Spanish elections and analyzed sentiment to classify the political alignment of bots and humans.

To sum up, as shown in Table \ref{tab:comparative_analysis}, most analyzed studies have focused on public reactions in English within a specific event. In contrast, this study is unique in its approach by analyzing topics, sentiments, and the prevalence of hate in $348,123$ public reactions to a range of news articles published by major Spanish media outlets in January 2021. The dataset includes reactions collected from two platforms: the web portals of the media outlets and Twitter. To our knowledge, this dual-platform collection of the same media content is the only study of its kind. By examining official news content from the top Spanish mass digital media, this study enables a horizontal analysis of the news and audience ecosystem, allowing us to understand the dynamics of issues, sentiment, and hate in cyberspace in a cross-cutting manner rather than being limited to a single event or news item.

\begin{table}[ht!]
\centering
\begin{tabular}{p{2.6cm}|c|c|l|l|l|p{1.4cm}|c|c|c}
\textbf{Work} & \textbf{Size} & \textbf{Social media} & \textbf{Web} & \textbf{Context} & \textbf{Lang.} & \textbf{Content type} & \textbf{Topics} & \textbf{Sent.} & \textbf{Hate} \\ 
\hline
\cite{GARCIA2021107057} & \textbf{6.5M} & \greencheck & \xmark & Event (COVID-19) & EN, PT & Tweets & \greencheck & \greencheck & \xmark \\ \hline
\cite{Sattar2021} & 1.2M & \greencheck & \xmark & Event (Vaccines) & EN & Tweets & \xmark & \greencheck & \xmark \\ \hline
\cite{WanParis2021} & 133K & \greencheck & \xmark & Event (Siege) & EN & Tweets & \xmark & \greencheck & \xmark \\ \hline
\cite{Fiesler2018} & 775 & \xmark & \greencheck & Data privacy & EN & \textbf{Reactions to news} & \xmark & \greencheck & \xmark \\ \hline
\cite{MAQSA} & \textit{N/A} & \greencheck & \greencheck & \textbf{Cross-cutting} & EN & \textbf{Reactions to news} & \greencheck & \greencheck & \xmark \\ \hline
\cite{Goodwin2018848} & 1K & \xmark & \greencheck & Water Reuse & EN & \textbf{Reactions to news} & \xmark & \greencheck & \xmark \\ \hline
\cite{CHOI201750} & 3.9M & \xmark & \greencheck & Event (MERS) & KO & Reactions to event & \greencheck & \greencheck & \xmark \\ \hline
\cite{CHEN20151780} & 1.6K & \greencheck & \greencheck & Event (Vaccines) & ZH & Reactions to event & \xmark & \greencheck & \xmark \\ \hline
\cite{Miranda2023} & 6.3M & \greencheck & \xmark & Event (COVID-19) & \textbf{ES} & Tweets & \greencheck & \greencheck & \xmark \\ \hline
\cite{9226407} & 5.8M & \greencheck & \xmark & Event (Elections) & \textbf{ES} & Tweets & \xmark & \greencheck & \xmark \\ \hline
Our & 348K & \greencheck & \greencheck & \textbf{Cross-cutting} & \textbf{ES} & \textbf{Reactions to news} & \greencheck & \greencheck & \greencheck \\
\hline
\end{tabular}
\caption{Comparison of public reaction studies detailing dataset size, social and web media types, contexts, languages, content types, and analyses conducted.}
\label{tab:comparative_analysis}
\end{table}

\section{Research methodology}\label{met}

\begin{figure*}[ht!]
	\centering{
		\includegraphics[width=0.95\textwidth]{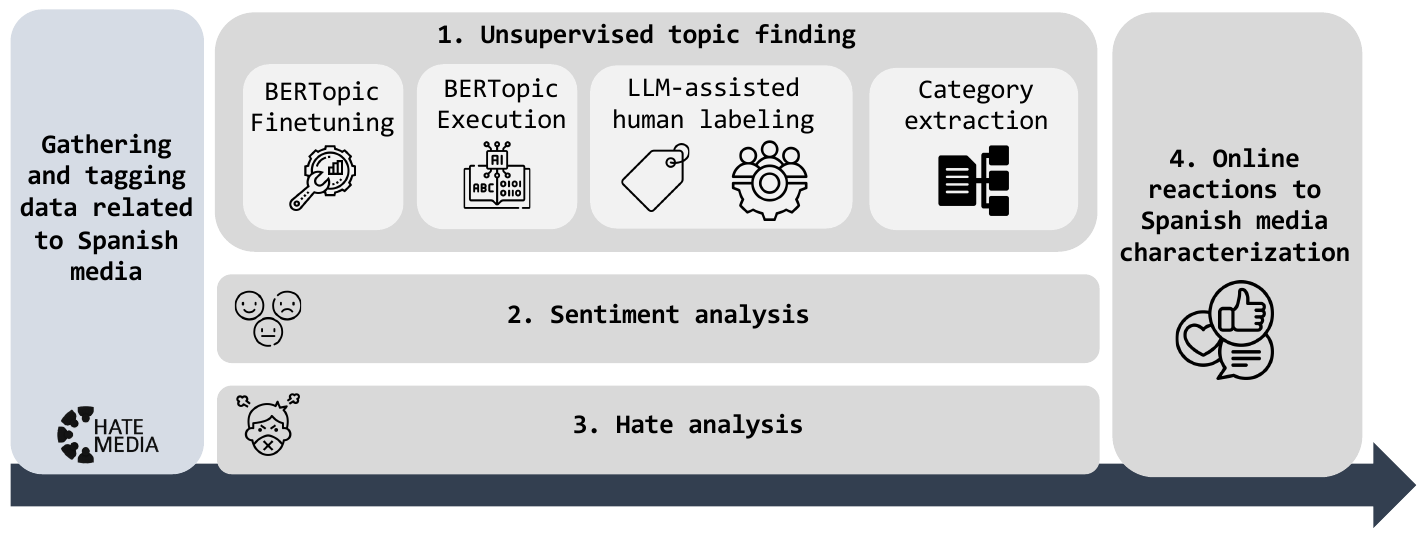}
	\caption{Methodology followed to gather and label the raw data, extract topics, analyze sentiments, analyze hate, and characterize Spanish media reactions.}\label{fig:met-diagram}
 }
	
\end{figure*}

The methodology followed encompasses four key stages: (i) gathering and tagging data related to Spanish media, (ii) sentiment analysis, (iii) topic finding, and (iv) online reactions to Spanish media characterization. Figure \ref{fig:met-diagram} represents the entire methodology process.

\subsection{Description of the dataset}

The dataset utilized in this study consists of posts and comments derived from Spanish media content gathered from the Hatemedia project \footnote{\url{https://www.hatemedia.es/metodologia-hatemedia/}}.

The data-gathering process occurred during the 31 days of January 2021 using automated web scraping techniques for $\mathbb{X}$ (previously Twitter) and institutional digital media portals (\textit{La Vanguardia}, \textit{ABC}, \textit{El Pa\'is}, \textit{El Mundo}, and \textit{20 Minutos}), to extract comments from readers on the news posted on the web and the other hand, replies to tweets where such news are advertised in $\mathbb{X}$. These public reactions derived from news content from these five Spanish media outlets were deduplicated and saved for tagging purposes. As a result, $193,066$ ($55.46\%$) online messages were collected from websites and $155,057$ ($44.54\%$) reactions were collected from $\mathbb{X}$ (tweets).

Using these gathered data, the tagging of all data took place between February and April 2022. The labeling process focused on identifying the presence of hate speech and categorizing it into specific types: religious, xenophobia, racist, misogyny, sexual, ideological, and other~\cite{DeLucasVicente2022}. Finally, the labeling was validated and reviewed collaboratively by the entire team.

Table \ref{tab:sample} presents a sample of the resulting dataset, which consists of six columns: `ID' (a unique identifier for each entry), `MEDIA' (the media outlet that posted the news), `PLATFORM' (either $\mathbb{X}$ or the media's website), `URL' (a link to the original content), `CONTENT' (the reaction to the news), and `HATE TYPE' (the type of hate expressed in the reaction).

Examining this sample reveals a reaction related to victims of terrorism news published by ABC on their website. This reaction criticizes Spanish politics and is categorized as ideological hate. Additionally, a satirical reaction to news related to the Spanish healthcare system during the pandemic, published by 20 Minutos, contains a joke about a patient's life expectancy and is categorized as non-hate. Lastly, a tweet derived from news related to the COVID-19 restrictions in Spain is categorized as an ``other'' type of hate, containing the expression ``\texttt{What a disaster of a country}.''

\begin{table*}[ht!]
\centering
\begin{tabular}{m{0.7cm}|m{1.2cm}|m{2cm}|m{4cm}|m{5cm}|m{2cm}}
\textbf{ID} & \textbf{MEDIA} & \textbf{PLATFORM} & \textbf{URL} & \textbf{CONTENT (translated)} & \textbf{HATE TYPE} \\
\hline
\hline
70283 & ABC & WEB & \href{https://www.abc.es/espana/abci-victimas-terrorismo-sin-victimas-pero-bildu-202106271316_noticia.html}{https://www.abc.es/ espana/abci-victimas-terrorismo-sin-victimas-pero-bildu-202106271316 \_noticia.html}  & Mediocre, this is the only country in the world that, in its rancid sectarianism, has managed to divide even the associations of victims of terrorism.
 & Ideological \\
\hline
61132 & 20MIN & WEB & \href{https://www.20minutos.es/noticia/4546587/0/enfermeras-espanolas-manifiesta-sintomas-ansiedad-pandemia/}{https:// www.20minutos .es/ noticia/4546587/0/ enfermeras-espanolas-manifiesta-sintomas-ansiedad-pandemia/}  & - Doctor, doctor, how is my analysis? How much time do I have left, doctor? - Well, I'd say about 10... - 10?, but 10 what... years, months, days...? - 9, 8, 7, 6...  & \textit{N/A}\\
\hline
6998 & 20MIN & TWITTER & \href{https://x.com/clipcinco/status/1346802795534540802}{https://x.com/ clipcinco/ status/ 13468027955 34540802} & What a disaster of a country & Other \\
\hline
\hline
\end{tabular}
\caption{Sample of the dataset used in this study with three reactions and its associated data.}
\label{tab:sample}
\end{table*}

\subsection{Unsupervised topic finding}

Initially, topics were extracted to address the first objective (O1). This step aligns with the initial phase of our methodology, which aims to identify the topics within public reactions to Spanish media content, excluding the prior description of the dataset, which will be applied in Section \ref{results:topics}.

Various methods exist for identifying the primary topics within a textual corpus, with one of the most widely used and effective being the probabilistic model known as Latent Dirichlet Allocation (LDA) \cite{fisher1936linear}. In LDA, a distribution of topics typically represents each document, and a distribution of words characterizes each topic. Despite its widespread use, LDA's effectiveness in analyzing social media data, such as Twitter, has faced significant criticism \cite{Egger2022}.

To address this issue, text embedding techniques have rapidly become popular in the natural language processing (NLP) area \cite{DBLP:journals/corr/MikolovSCCD13}. Among these, BERTopic \cite{grootendorst2022bertopic} enhances the cluster embedding approach by leveraging state-of-the-art language models. In particular, BERTopic operates through several phases: first, document embeddings are created using a pre-trained language model like BERT. Following this, TF-IDF is applied to create key term vectors. Subsequently, UMAP is used for dimensionality reduction. Finally, clustering algorithms such as HDBSCAN group similar documents into distinct topics. This approach positions BERTopic as a leading model for topic modeling~\cite{PASTORGALINDO202467}.

In a preliminary study~\cite{JNIC24}, the performance of LDA and BERTopic was compared, using the coherence measures in three different random samples to evaluate both models. Specifically, the $C_v$ coherence measure was employed as the evaluation metric, which combines the indirect cosine measure with normalized pointwise mutual information and the boolean sliding window \cite{10.1145/2684822.2685324}. Higher coherence scores indicated more meaningful and interpretable topics. As a result of this evaluation, the BERTopic model achieved a higher coherence value than LDA, demonstrating superior performance for our use case.

\subsubsection{BERTopic finetuning}\label{bertopic-met}

Before applying BERTopic, it was necessary to determine the optimal number of topics for the entire dataset. This involved fine-tuning the hyperparameters to enhance the consistency and accuracy of the BERTopic models. The following hyperparameters were specifically adjusted:

\begin{itemize}
    \item \textit{n\_neighbors}: This parameter defines the number of neighbors used in constructing the UMAP representation space. A range of $5$ to $25$, in increments of five, was tested to balance computational efficiency with the need for a smooth representation that aids pattern identification.

    \item \textit{min\_cluster\_size}: This parameter sets the minimum cluster size in HDBSCAN. A range of $50$ to $600$, in increments of $50$, was evaluated to identify meaningful clusters without merging distinct topics into excessively large clusters.
\end{itemize}

The coherence value ($C_v$) and the number of topics in each combination were analyzed. Significant variation in the number of topics and coherence is observed concerning the values of the hyperparameters, with a $C_v$ maximum value of $0.74$ and a minimum of $0.47$.

The configuration with the best performance (max coherence and a minimum number of topics) is the combination of $81$ topics having a $C_v$ of $0.74$, with $20$ \textit{n\_neighbors} and \textit{min\_cluster\_size} of $350$. This choice can lead to better interpretability and ease of analysis without compromising the quality of topic coherence.

\subsubsection{BERTopic execution}\label{bertopic-exec}

Once the optimal hyperparameters were determined through fine-tuning, the BERTopic model was executed to extract topics from the Spanish public reactions. In particular, the BERTopic model was configured with the ``multilingual'' language parameter to accommodate our Spanish dataset. Moreover, Spanish stopwords were removed to filter out common words that do not contribute to the topic identification process.

Furthermore, for dimensionality reduction, the UMAP model was initialized with seven components, a minimum distance of $0.1$, and cosine as the distance metric to balance computational efficiency and smooth data representation. Moreover, the HDBSCAN model used Euclidean distance and the `eom' method for cluster selection to identify meaningful clusters without merging distinct topics into excessively large clusters.

Finally, when using HDBSCAN, some outlier documents were created that do not fall within any created topics. These are labeled as cluster -1. By minimizing these outliers, the resulting clusters will better represent the underlying structure of the data, ensuring that each document is meaningfully associated with a specific topic and enhancing the accuracy and reliability of the topic modeling process. For this purpose, a function provided by BERTopic (\texttt{reduce\_outliers}) was employed to reduce outlier documents and label them as non-outlier topics using the c-TF-IDF \footnote{\url{https://maartengr.github.io/BERTopic/getting_started/outlier_reduction/outlier_reduction.html}}.

\subsubsection{LLM-assisted topic labeling}

The output of the BERTopic execution includes clusters of similar documents along with representative keywords. To facilitate the labeling task of the authors, four LLMs were utilized to suggest appropriate labels for the clusters based on their associated keywords and documents. For this task, LLAMA2 from Meta AI \cite{touvron2023llama}, Zephyr from Mistral \cite{rafailov2023direct}, GPT-3.5 from OpenAI \cite{gpt3.5}, and GPT-4o from OpenAI \cite{gpt4o}, launched in 2024, were employed due to their strong reputations and performance in natural language processing tasks. The authors reviewed the four outputs and tailored the final label collaboratively.

\paragraph{\textbf{Prompt engineering}}

Since prompts are crucial for text generation using LLMs, the prompt recommended by the BERTopic framework for labeling tasks was used\footnote{\url{https://maartengr.github.io/BERTopic/getting\_started/representation/llm\#llama-2}} and take into account main c-TF-IDF topic's keywords and ten most representative documents. However, the labels were not representative enough for the clusters. To mitigate this, the most representative posts were omitted to avoid introducing bias into the model. Instead, a sample of posts from each topic was provided to ensure a more diverse and comprehensive representation of the topics.

Despite these adjustments, the outputs from the LLMs were too specific. Given the large number of reactions with diverse topics within each cluster, a sample did not adequately represent the main topic. To address this issue, the most representative keywords from each cluster were extracted using KeyBERT, a BERT-based method that leverages pre-trained language models to identify the most relevant keywords for a reaction, and Maximal Marginal Relevance (MMR), an information retrieval technique that ensures the extracted keywords are not only relevant but also diverse. Combining these methods provided a nuanced, diverse and comprehensive set of keywords encapsulating each topic's core themes.

Following keyword extraction, the keywords were lemmatized to reduce words to their base or root form, ensuring that word variations are treated as a single entity. This step eliminated redundancy and ensured that similar concepts were grouped.

Using this approach, the outputs from the LLMs were more representative and useful in the labeling task. Therefore, Prompt \ref{llama-gpt} was employed for each cluster, omitting the posts and instructing the LLM to act as an expert assistant in labeling topics by providing the keywords of the topic's posts (text in blue).

\begin{lstlisting}[language=python, abovecaptionskip=4pt, captionpos=b, caption =Prompt used to generate labels of each topic with LLMs,label={llama-gpt}, linewidth=\columnwidth]
 You are a helpful, respectful and honest assistant for labeling topics.

 The topic is described by the following keywords: *\color{delim2}`\textbf{[KEYWORDS]}'*
 
 Based on the information above, please create a short label for this topic in
 English. Make sure you only return the short label and nothing more.
\end{lstlisting}

Thus, each of the four LLMs receives the same prompt and suggests its own label for each cluster (topic).

\paragraph{\textbf{Topic labeling}}

Due to the diverse nature of short messages within clusters and the presence of figurative and sarcastic in online content, the suggested labels were not directly usable and the authors refined them. Consequently, each of the three authors independently reviewed the four labels suggested by the LLMs for each topic. This independent review process allowed for an unbiased evaluation of the suggested labels. After reviewing these options, the authors selected the most fitting label, possibly refining or generalizing the label as needed to encapsulate the topic best. Finally, a majority vote reached a consensus, ensuring that most authors agreed upon the chosen label.

While human review and understanding of the world are crucial in ensuring the accuracy and relevance of topic labels, the assistance provided by LLMs offers a significant advantage over completely manual labeling by providing initial suggestions that facilitate and accelerate the creation of labels, making the process more efficient.

\subsubsection{Category extraction}\label{category-extraction}

This step's objective is to classify the $81$ identified topics into categories, facilitating a more manageable and meaningful analysis. To achieve this, the first step involves calculating the embeddings for the topics, a powerful tool for representing textual data in a dense vector space and capturing semantic similarities between topics. Four of the most widely used word embedding models were tested, including BERT~\cite{devlin2019bertpretrainingdeepbidirectional}, RoBERTa~\cite{liu2019robertarobustlyoptimizedbert}, FastText from Meta~\cite{joulin2016bagtricksefficienttext}, and GloVe~\cite{pennington-etal-2014-glove}. Among these, BERT produced the most accurate and meaningful embeddings for these topic labels, as validated through manual comparison with the outputs generated by other embedding techniques.

With the embeddings generated using BERT, a hierarchical clustering algorithm was employed to determine the optimal number of clusters. By applying a threshold of $1.6$ to the dendrogram generated by the hierarchical clustering, nine categories (clusters) were identified.

Once the optimal number of clusters was determined, the k-means algorithm was executed to refine the clustering process, ensuring cohesive and well-defined clusters. While hierarchical clustering provided a high-level understanding of the data's structure,  k-means helped to polish and clearly define these clusters.

This clustering task was challenging due to the variety of topics, and the three authors reviewed and adjusted the resulting clusters (categories) manually to ensure accuracy and relevance.

Finally, each cluster (category) was assigned a representative label. This labeling process involved analyzing the topics within each cluster to manually create concise and descriptive labels in consensus by the three authors. 

\subsection{Sentiment analysis}\label{sentiment-met}

The primary objective of sentiment analysis (O2) was to identify a text's emotional tone, opinion, subjectivity, and attitude. This research employed TweetNLP \cite{camacho-collados-etal-2022-tweetnlp} to categorize each post as having positive, negative, or neutral sentiment, utilizing advanced language models tailored for social media contexts. TweetNLP was selected due to its superior performance, as demonstrated by the authors using the ``Macro-F1'' evaluation metric, outperforming other options such as BERTweet \cite{bertweet} and TweetEval \cite{DBLP:journals/corr/abs-2010-12421}, ensuring more accurate and reliable sentiment analysis of public reactions derived from to news content from five Spanish media outlets.

\subsection{Hate analysis}\label{hate-met}

The primary objective of the hate analysis (O3) was to investigate the prevalence of hate speech within the manually labeled dataset in collaboration with the Hatemedia project. This analysis focused on identifying and examining hate speech across the topics and categories identified in public reactions to content from the Spanish media outlets included in this research.

To accurately detect and categorize hate speech, a systematic methodology was employed. The process began with a clear definition of hate speech, aligned with international standards, focusing on content that incites violence, discrimination, or hostility against individuals or groups based on characteristics such as race, religion, gender, or political affiliation. A team of trained annotators manually reviewed a representative sample of comments from readers collected from $\mathbb{X}$ (formerly Twitter) and the websites of five major Spanish media outlets: \textit{La Vanguardia}, \textit{ABC}, \textit{El Pa\'is}, \textit{El Mundo}, and \textit{20 Minutos}.

Annotators assigned labels to each message, categorizing them as hate speech, offensive but not hate speech, or non-offensive. To ensure consistency and reliability, multiple annotators independently reviewed each message. Disagreements were resolved through discussion and consensus-building processes. Additionally, regular meetings and calibration sessions were held to standardize criteria and minimize subjective bias among annotators.

\subsection{Online reactions to Spanish media characterization}\label{met:topic-media}

The primary objective of characterizing the five Spanish media outlets (O4) was to examine the differences in topic distribution, sentiment, and hate speech levels caused across each Spanish media included in the analysis (\textit{La Vanguardia}, \textit{ABC}, \textit{El Pa\'is}, \textit{El Mundo}, and \textit{20 Minutos}).

\subsubsection{Differences between topics, sentiment and hate speech}

Concerning topics, due to the large number of clusters, specifically $81$ distinct topics, a comprehensive approach was necessary to discern differences in the topics discussed in reactions to each media outlet. The goal was to understand how the reactions varied across different platforms regarding the topics discussed. This approach involved several phases:

\begin{enumerate}
    \item Normalization: Due to the differing number of documents from each media outlet ($i$), normalized values are used to ensure comparability and accuracy. Normalization is achieved by dividing each media outlet and topic ($t$), the document count ($D_{ti}$) by the total number of documents for the respective media outlet, mitigating potential biases arising from unequal document volumes.

    This normalized value is denoted as shown in formula \ref{formula:normalization}.

 \begin{equation}
 ND_{ti} = \frac{D_{ti}}{\sum_{t=1}^{Topics} D_{ti}}
 \label{formula:normalization}
 \end{equation}    

    \item Average number of messages for each topic: For each topic ($t$), the mean distribution ($M_t$) across all media outlets is calculated. This mean serves as a benchmark for comparing individual media outlets' coverage of that topic.

   \item Media outlet's topic coverage deviation: For each media outlet ($i$) and topic ($t$), the deviation is calculated by subtracting the mean distribution from the normalized document count. Consequently, the formula for the deviation is as follows: $ND_{ti} - M_t$. This deviation indicates how much a specific media outlet's topic coverage diverges from the average coverage.

\end{enumerate}

These steps collectively provide a structured approach to identifying and quantifying the differences in topic discussions in public reactions to Spanish media content.

Finally, the distribution of sentiments (positive, neutral, and negative) and hate content (no, yes) caused across Spanish media outlets were analyzed. For this purpose, the dataset was grouped by media outlet to obtain the public reactions of each type for each media outlet. This approach facilitates the examination of differences in sentiments and hate speech levels across each Spanish media included in the analysis.

\subsubsection{Differences between Twitter and news web portal's public reactions}

The public reactions to news across web portals and $\mathbb{X}$ (Twitter) can differ. While the news in web portals is daily and contrasted, $\mathbb{X}$ provides a dynamic and real-time communication channel often characterized by brevity and immediacy. Therefore, the news published on this social network will be designed for interaction. Consequently, the nature and tone of public reactions on these platforms will likely diverge significantly.

For this reason, to quantify the statistical differences between the topics discussed on $\mathbb{X}$ and those on news web portals, the distinct characteristics of each platform's topics, sentiments and hate were analyzed to evaluate whether the public discourse is different on $\mathbb{X}$ than on news web portals. This analytical approach discerns these differences before characterizing the public reactions of each platform.

First, the dataset was split into two parts: content originating from web sources and content from $\mathbb{X}$. Following this division, the topics, sentiment, and hate speech were analyzed separately for each subset to determine if significant differences exist between the public reactions across these platforms.

The distribution of reactions along topics was considered a variable, contrasting sentiments (positive, negative, or neutral) and hate (presence of hate or not). For this reason, a different approach was employed.

In particular, for topic distribution, the Shapiro-Wilk test \cite{ShapiroWilk1965} was used to evaluate normality, having extremely low p-values ($1.425 \cdot 10^{-19}$ for $\mathbb{X}$ and $1.485 \cdot 10^{-19}$ for web portals), indicating non-normal distribution. Consequently, the Mann-Whitney U test \cite{doi:https://doi.org/10.1002/9780470479216.corpsy0524} was applied, resulting in a p-value of $0.614$, suggesting no significant differences between the platforms regarding topic distribution.

For sentiments, Levene's test \cite{Levene_1960} confirmed variance homogeneity across negative, neutral, and positive sentiments, with p-values of $0.576$, $0.297$, and $0.794$, respectively. This allowed for a MANOVA test \cite{doi:https://doi.org/10.1002/9781118445112.stat02476}, which identified statistically significant differences in sentiment distribution between the platforms. However, Cohen's d \cite{SullivanFeinn2012} values revealed small effect sizes, indicating minimal practical significance.

In the analysis of hate content, Levene's test again confirmed equal variances, with p-values of $0.245$ for hate and $0.967$ for non-hate content. MANOVA results showed statistically significant differences in hate distribution across platforms. Nonetheless, Cohen's d indicated small effect sizes, suggesting that while there are differences in hate content between platforms, these differences are not practically significant. Manual human verification confirmed that the differences in content between platforms were minimal.



\section{Experimental results}\label{results}

In this section, the full methodology is applied to the dataset to analyze public reactions to Spanish media content. This section presents the findings of the topic modeling, sentiment analysis, and hate speech analysis. The code used in this study has been made available on GitHub\footnote{\url{https://github.com/Alexbl7/Toxicity-in-Spanish-information-environment}}.

\subsection{\textbf{Objective 1: Topics of public reactions}}\label{results:topics}

First, topics are extracted to address the first objective (O1). As a result of applying the BERTopic model to our dataset, described in Section \ref{bertopic-met}, $81$ clusters (topics) were obtained. The methodology considered $1,864$ posts ($0.54\%$ of the total) as outliers and $8,452$ posts ($2.43\%$) as irrelevant content or low quality. Therefore, $10,316$ posts ($2.96\%$ of the total) were discarded, leaving $337,807$ posts ($97.04\%$) for further analysis as the base for the rest of the article.

\subsubsection{Distribution of topics}

The resulting clusters, along with the sample of reactions, were prompted to the four LLMs to propose label candidates. The authors then selected and refined these proposals. The labels for each topic, the outputs from the LLMs, and the authors' final decisions are available online\footnote{\url{https://osf.io/5yncw?view_only=713b55b32e764af6a8fb4eb8170ee3af}}. 

Table \ref{tab:bertopic_clusters} presents the top 10 topics identified in the dataset using BERTopic that generated the most reactions along their cluster associated by BERTopic, the number and percentage of associated posts, the KEYBERT keywords and the LLM-assisted human label.

The conversation expressions topic, with 26,681 posts (7.9\% of the total), is the topic with the most public reactions associated, including casual conversational phrases, such as ``\texttt{What's up?}'' or ``\texttt{okay,  perfect}.'' In this vein, the negative expressions cluster follows closely, indicating the prevalent use of negative language in public reactions, with 25,698 posts (7.61\%) containing posts like ``\texttt{Nobody does anything}'' or ``\texttt{It will amount to nothing}''.

Regarding social issues, the gender \& feminism topic, with 15,880 posts (4.7\%), shows significant engagement with gender issues. Examples include reactions like ``\texttt{it's going to be pretty funny the mess this law is going to make that I'm not good enough to compete against other men because I declare myself a woman}.''

Moreover, the vaccination topic (13,019 posts, 3.85\%) reflects the public interest in COVID-19 vaccination debates. Examples highlight public reactions, such as concern over misinformation and vaccine skepticism. For instance, one post discusses the spread of fake news: ``\texttt{The forensic chief in Germany was misquoted to create fake news. One person’s claim cannot stand against the whole scientific community.}'' Another post expresses distrust in government officials: ``\texttt{Does anyone believe this minister has not already been vaccinated, given how dishonest the government is?}.''

In the Spanish context, the Madrid cluster (11,424 posts, 3.38\%) focuses on regional concerns, with posts such as ``\texttt{If more than 6 million people live in Madrid and the damages are in the order of millions of euros, they come out to a few euros per person, which does not seem a catastrophe.}.''

The third topic containing expressions in the top 10 topics with more reactions is the time expressions topic (11,222 posts, 3.32\%), which includes posts expressing time-related concerns and frustrations, such as ``texttt{I'm late, again}'' and ``\texttt{how much longer are we going to wait?}.''

Concerning public administration, topics like the justice system (9,390 posts, 2.78\%) often include posts such as ``\texttt{justice is too slow}'' or ``\texttt{the corrupt always get away with it}'', reflecting public discontent with perceived inefficiencies and corruption within the judicial system.

Moreover, the media cluster (9,360 posts, 2.77\%) includes criticism of media practices, with posts like ``\texttt{It should take journalism studies to stand in front of a television camera}'' and ``\texttt{media dictatorship is an understatement},'' highlighting skepticism towards news reporting.

Furthermore, the municipal funding topic (8,788 posts, 2.6\%) covers concerns about local budgets, exemplified by posts such as ``\texttt{Why is our city spending millions on a new stadium when our schools are underfunded?}'' or ``\texttt{Our community centers are closing due to 'budget constraints,' yet there's always money for new police equipment. This is unacceptable!}.''

Lastly, the Spanish politics topic (8,641 posts, 2.56\%) features a range of opinions on national political issues, with comments like ``\texttt{These puppies of the PP, the alpha males of the Spanish state, are the masters of the universe}'' and ``\texttt{Spanish politics is a circus},'' reflecting public discontent.

\begin{table*}[t!]
\centering
\begin{tabular}{p{1.5cm}|p{1.5cm}|p{1.6cm}|p{7.3cm}|p{2.1cm}}

\textbf{BERTopic cluster} & \textbf{Number of posts} & \textbf{Percentage of posts} & \textbf{KEYBERT keywords} & \textbf{LLM-assisted human label} \\
\hline
\hline
1 & 26,681 & 7.9\% & pasa, vas, vale, hace, si, ser, ve, pues & Conversation expressions \\
\hline
8 & 25,698 & 7.61\% & nadie, ning\'un, nunca, tampoco, in\'util & Negative expressions \\
\hline
2 & 15,880 & 4.7\% & feminista, mujer, feminismo, mujeres, feministas & Gender \& feminism \\
\hline
0 & 13,019 & 3.85\% & vacunaciones, vacunados, vacunar, vacunaci\'on, vacunarse & Vaccination \\
\hline
3 & 11,424 & 3.38\% & madrid, madrile\~{n}a, madrile\~{n}o, madrile\~{n}os, espa\~{n}a & Madrid \\
\hline
11 & 11,222 & 3.32\% & tarde, retrasados, retraso, a\'un, tardando & Time expressions \\
\hline
6 & 9,390 & 2.78\% & condenados, presos, c\'arcel, delincuentes, condenado & Justice system \\
\hline
5 & 9,360 & 2.77\% & noticia, noticias, news, period\'istica, period\'istico & Media \\
\hline
15 & 8,788 & 2.6\% & proyecto, proyectos, sostenible, presupuesto, infraestructuras & Municipal funding\\
\hline
4 & 8,641 & 2.56\% & espa\~{n}oles, espa\~{n}olitos, s\'anchez, espa\~{n}ola, espa\~{n}ol & Spanish politics \\
\hline 
\hline 
\end{tabular}
\caption{Top 10 topics with the highest number of reactions identified in the dataset using BERTopic, along with corresponding post counts, percentages, KEYBERT keywords, and LLM-assisted human labels.}

\label{tab:bertopic_clusters}
\end{table*}

Notably, these topics are highly fine-grained, more informative and less general than traditional categorization results. For instance, narratives about particular political parties, mentions of specific leaders, and discussions on certain health are clearly identified.

\begin{table}[h!]
\centering
\footnotesize
\begin{tabular}{>{\columncolor{green!10!white}}p{1.8cm}|>{\columncolor{blue!10!white}}p{1.5cm}|>{\columncolor{black!70!white}\color{white}}p{1.6cm}|>{\columncolor{orange!10!white}}p{1.4cm}|>{\columncolor{violet!11!white}}p{1.2cm}|>{\columncolor{red!10!white}}p{1.4cm}|>{\columncolor{brown!10!white}}p{1.8cm}|>{\columncolor{pink!25!white}}p{1.7cm}|>{\columncolor{gray!11!white}}p{1.5cm}}

\textbf{Social issues (21)} & \textbf{Expressions \& slang (6)} & \textbf{Political issues (14)} & \textbf{Public administration (8)} & \textbf{Spanish politics (6)} & \textbf{COVID-19 (4)} & \textbf{International relations \& immigration (10)} & \textbf{Hobbies \& lifestyle (7)} & \textbf{Safety \& Primary Sectors (5)} \\
\hline
\hline
Gender \& feminism & Conversation expressions & US politics & Justice system & Madrid & Vaccination & Latin America & Food & Seamanship \\
\hline
Death \& violence & Negative expressions & Socialism \& communism & Healthcare system & Spanish politics & COVID-19 contagion & China & Music & Energy sector \\
\hline
Winter climate & Time expressions & Government & Electoral system & Catalonia & Pandemic management & European Union & Reading \& writing & Military \\
\hline
Money matters & Insults \& slurs & Political system & Municipal funding & Pablo Iglesias & COVID incidence & Homeland & Beverages & Law enforcement \\
\hline
Racism & Sarcastic expressions & Fascism & Education system & Salvador Illa & & Russia & Astronomy & Emergency services \\
\hline
Lies & Internet slang & Government ministries & Taxation & Spanish political parties & & Immigration & Photography & \\
\hline
Animals & & Royalty \& monarchy & Pension system & & & Global issues & Cinematography & \\
\hline
Hospitality \& tourism & & Amnesty & Transport system & & & Moroccan immigration & & \\
\hline
Vehicles \& street & & Democracy & & & & Currency & & \\
\hline
Drugs \& pharmaceuticals & & Terrorism & & & & France & & \\
\hline
Media & & Dictatorships & & & & & & \\
\hline
Religion & & Left \& right wing & & & & & & \\
\hline
Humor \& laughter & & Populism & & & & & & \\
\hline
Airlines & & Independentism & & & & & & \\
\hline
Social media & & & & & & & & \\
\hline
Snowy weather & & & & & & & & \\
\hline
Islam & & & & & & & & \\
\hline
Mental health & & & & & & & & \\
\hline
Television & & & & & & & & \\
\hline
Smoking & & & & & & & & \\
\hline
Cycling and motorcycling & & & & & & & & \\
\hline
\hline 
\end{tabular}
\caption{Topics detected in public reactions to Spanish media across different categories.}
\label{tab:clusters}
\end{table}

\subsubsection{Extracted categories}

Following the methodology stated in Section \ref{category-extraction}, we classify the $81$ identified topics into categories, and the embeddings for the topics were calculated. Using these embeddings, PCA was applied for dimensionality reduction; the resulting clusters were manually validated and labeled by the three authors.

As shown in Table \ref{tab:clusters}, the resulting $81$ labeled topics are grouped into nine categories of groups:

i) \social{Social issues}: is the most popular category, with $75,048$ reactions ($22.22\%$), and the most diverse, with $21$ topics assigned, considering aspects such as gender and feminism, deceit, media criticism, and mental health. It includes frustration, humor, and sarcasm with the aforementioned social issues. Having post like ``\texttt{find yourself a psychologist to bring you back to reality}'' or ``\texttt{and neither God nor anyone else will change this}.''

ii) \expressions{Expressions \& slang}: is the second most populated group, with $68,741$ posts (20.35\%) encompassing a diverse range of idioms, informal sentences, and sarcastic references as well as insults and slurs, and internet slang such as ``\texttt{XD},'' ``\texttt{jaja}'' or ``\texttt{in the end they all say the same thing}.'' This category is not centered around a specific theme. Instead, it captures the wide array of casual and spontaneous expressions found in online interactions, showcasing the reality of contemporary social media communication.

iii) \politic{Political issues}: with $39,851$ posts (11.80\%), is the second most diverse, having 14 topics assigned. It covers content related to political systems, governance, economic ideologies, public policies, and societal structures, encapsulating broad discussions on governmental functions, political movements, and ideological debates. With posts, such as ``\texttt{real socialism has not worked in any country}'', ``\texttt{politicians always promise gold and silver, but then leave us empty-handed}'' or ``\texttt{Corruption is the cancer of our politics}''.

iv) \public{Public administration}: groups $39,505$ reactions ($11.69\%$) in eight topics, encompassing messages related to the hospital healthcare system, education system, emergency services, social services, or transport system, among others. Having posts such as ``\texttt{we used to say that we had the best health care in the world, but now we see that we did not, but we already saw it when the hospital corridors were overcrowded year after year with the flu epidemic},'' ``\texttt{the firefighters and emergency services are to be thanked for their work}'' or ``\texttt{they don't care if people are cramped in public transport}.''

v) \spain{Spanish politics}: with $37,375$ reactions (11.06\%) in six topics, including discussions on the independence movement in Catalonia, Spanish politics, ministerial, political parties, and also content related to Spanish politicians such as Pablo Iglesias (leader of the Spanish left-wing party `Podemos') and Salvador Illa (Spanish minister of health during the pandemic era). With posts like ``\texttt{podemos no longer publicly hides what was becoming evident: that it works to eliminate capitalist property and guarantee the means of subsistence}'' or ``\texttt{more than ever, Vox for the good of Spain}.''

It should be noted that although there is this specific category of Spanish politics, the entire dataset is in the Spanish context. Therefore, most comments in other categories are also biased toward the Spanish domain.

vi) \covid{COVID-19}: this category is clearly defined. It includes topics such as vaccination efforts, the spread of COVID-19, and case numbers and mortality rates through four topics in $25,320$ posts ($7.50\%$), being the least diverse. Posts such as ``\texttt{Do we live the rest of our years locked up in fear of a virus with a standard fatality rate like the common flu? Forever? Do we never leave the house again?}'' or ``\texttt{What does it matter how many infected there are if no data on the dead is provided}.''

vii) \inter{International relations \& immigration}: grouping with $24,488$ posts (7.25\%) in $10$ topics, including discussions on countries like Venezuela, China, Russia, and immigration. With posts such as ``\texttt{The Chinese regime has no right, much less to criticize or say anything, while they don't meet the minimum human rights standards}'', ``\texttt{I remember all those pundits who predicted hell for the UK outside the EU}'' or ``\texttt{Morocco as an Islamic society is a social failure I could tell you about because I studied it.}.''

viii) \media{Hobbies \& lifestyle}: is less diverse, with seven topics, and fewer popular with $15,492$ posts (4.58\%), covering various topics related to everyday life and entertainment. It includes music, photography, food, and astronomy. Having posts like ``\texttt{All musicians and songwriters know that if there is a more recurring theme than love for writing songs, it is heartbreak, especially when it involves personal experiences},'' ``\texttt{Better old-school food than stupid and pretentious menus from insecure modernists}'' or ``\texttt{Yesterday's photos look like they were taken from the movie Batman}.''

ix) \transport{Safety \& primary sectors}: groups $11,987$ posts (3.55\%) in five topics, encompassing the energy sector, military, emergency services, and law enforcement topics. With posts such as ``\texttt{This is the army we all pay for and that serves only a few}'' or ``\texttt{The rise in gas prices is not trivial, as it is one of the factors driving up the value of energy the most}.''

\vspace{0.3cm}

\paragraph{\textbf{Distribution of categories}}

\begin{figure*}[t!]
	\centering{
		\includegraphics[width=\textwidth]{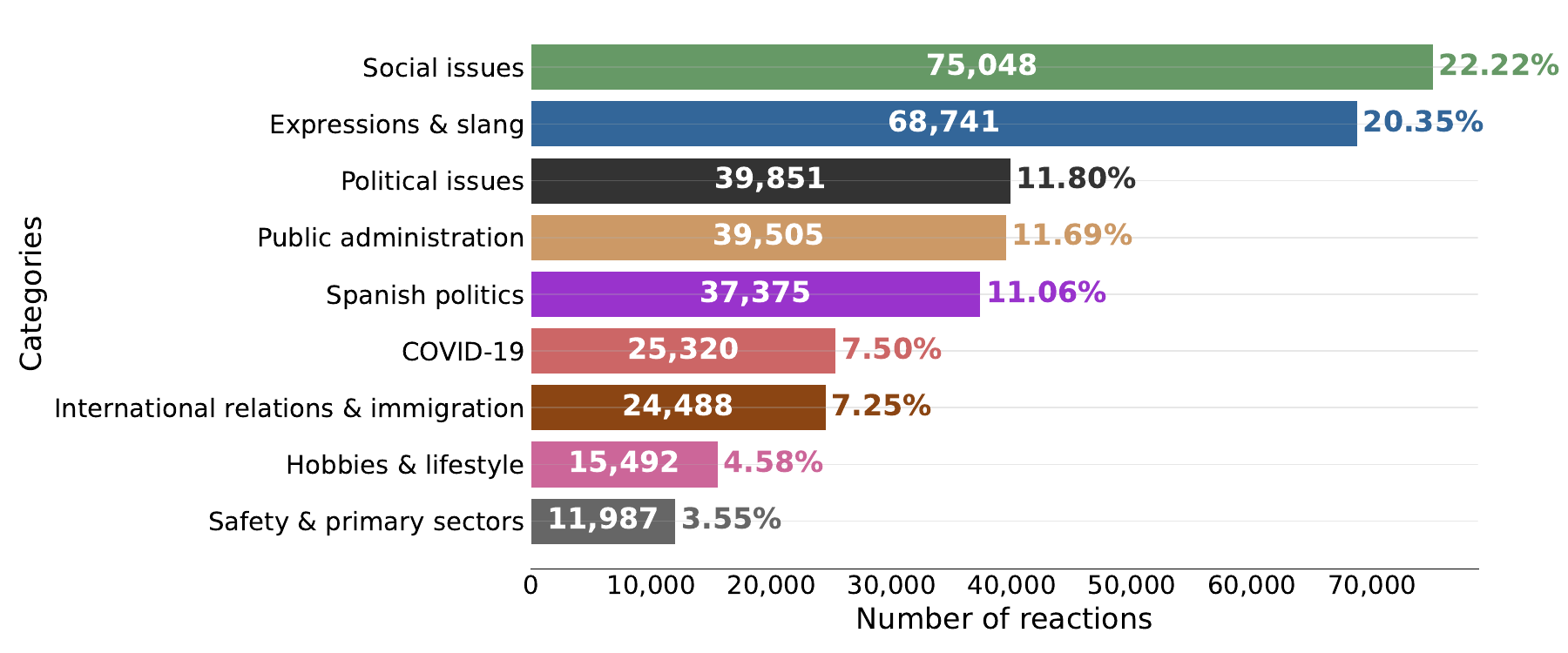}
	\caption{Distribution of public reactions caused by Spanish media content per category.}\label{fig:distribution-cat}
 }
\end{figure*}

As reflected in Table~\ref{tab:clusters}, the categories are unbalanced regarding topics covered. In summary, Figure \ref{fig:distribution-cat} illustrates the distribution of public reactions per category. 

The most popular category is \social{Social issues} with 75,048 reactions (22.22\%), followed by \expressions{Expressions \& slang} with 68,741 reactions (20.35\%). Other categories include \politic{Political issues} with 39,851 reactions (11.80\%), \public{Public administration} with 39,505 reactions (11.69\%), and \spain{Spanish politics} with 37,375 reactions (11.06\%). Additional categories are \covid{COVID-19} with 25,320 reactions (7.50\%), \inter{International relations \& immigration} with 24,488 reactions (7.25\%), \media{Hobbies \& lifestyle} with 15,492 reactions (4.58\%), and \transport{Safety \& primary sectors} with 11,987 reactions (3.55\%).

\vspace{0.3cm}
\paragraph{\textbf{Visualization of categories}}

\begin{figure*}[t!]
	\centering{
		\includegraphics[width=\textwidth]{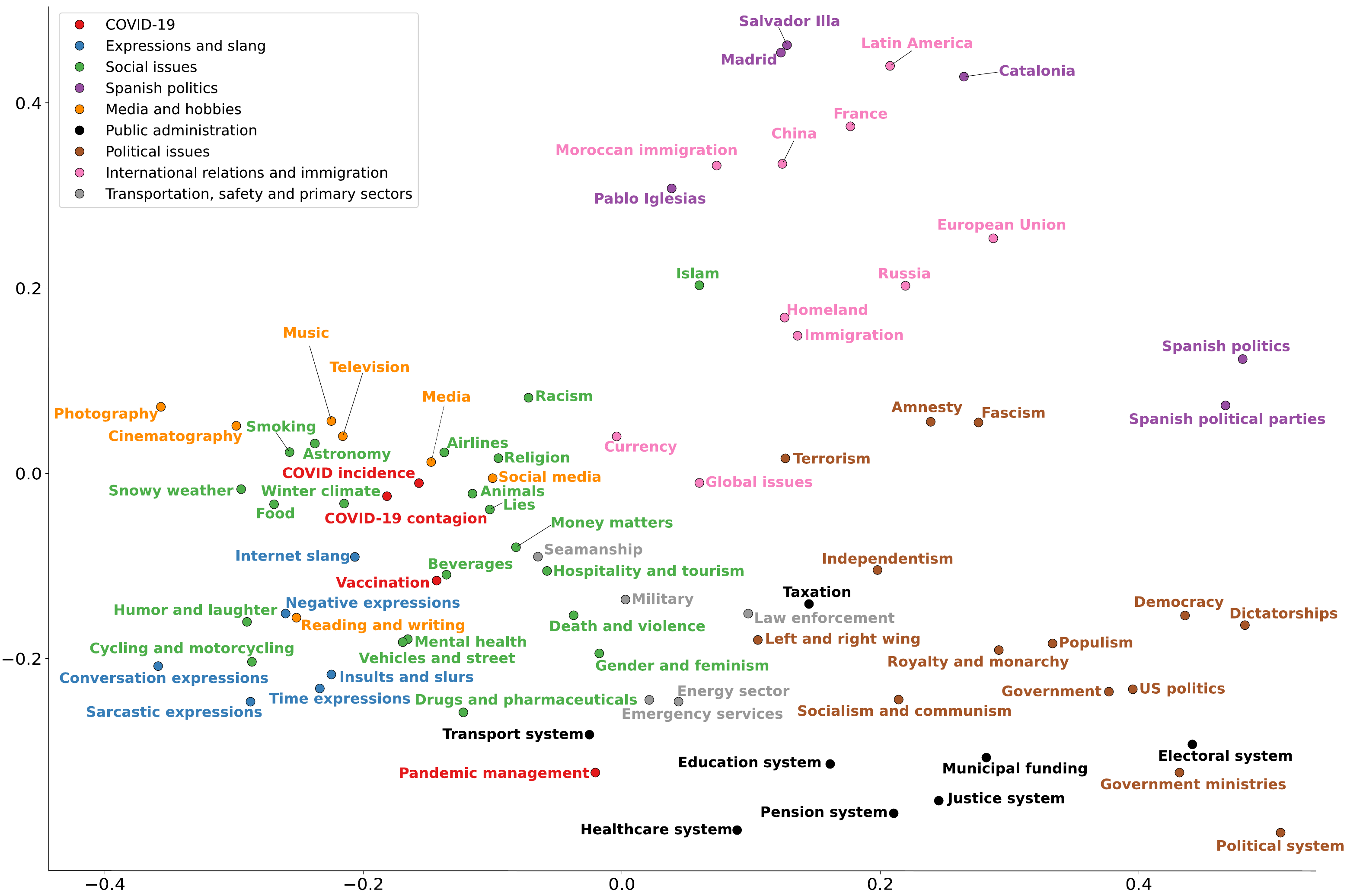}
	\caption{Resulting clusters (colored categories) of topics identified in public reaction around Spanish media using PCA for dimensionality reduction.}\label{fig:clusters}
 }
 
\end{figure*}

In such a scenario, Figure \ref{fig:clusters} illustrates the resulting clusters of categories with topics identified in public reactions around Spanish media, using PCA for dimensionality reduction. The scatter plot represents various topics, color-coded according to their respective categories. This color scheme will be used in the rest of the paper to identify these categories. 

Each point represents a specific topic, and the clusters indicate how closely related these topics are in the public discourse. Certain topics within clusters are positioned very close, suggesting a strong relationship or overlap in public perception. For example, within the \politic{Political issues} cluster, topics like \politic{US politics}, \politic{Socialism \& communism}, \politic{Government}, and \politic{Political system} are tightly grouped, reflecting their interconnected nature in public discussions.

In this context, certain topics that appear very close do not belong to the same category, and others very far away do. This is due, firstly, to the fact that this is a 2D visualization (PCA) of a broader and more complex dimensional space and, secondly, to the fact that the categories are reviewed and modified, if applicable, by the article's authors.

In particular, very social and political topics discuss the reality of January 2021. These discussions indicate a continued focus on governance, public health, and economic issues, echoing the concerns noted in the introduction regarding 2021 events, a period heavily influenced by the social, economic, and political impacts of the COVID-19 pandemic, as well as the devastating effects of the Filomena storm in Spain.

\subsection{\textbf{Objective 2: Sentiment analysis}}\label{results:sents}

Following the methodology stated in Section \ref{sentiment-met}, a sentiment analysis procedure with \textit{TweetNLP} is applied to the public reactions to Spanish media content. Table \ref{tab:all-results} aggregates, under the ``\textit{Sentiments}'' column, the number of reactions labeled with positive, neutral, and negative sentiment per category. In addition, the ``\textit{Hate}'' column indicates the number of reactions labeled as hateful content (will be discussed in Section \ref{results:hate}). The last column, ``\textit{Total}'', represents the total count of reactions for each category.

\subsubsection{Distribution of sentiments in the dataset}

The distribution of sentiments (positive, negative, and neutral) across $337,807$ public reactions considered by BERTopic is analyzed. The dataset has a notable prevalence of toxic content, with $211,567$ negative reactions ($62.63\%$). Posts such as ``\texttt{Madrid is a failed community, governed by two failed leaders},'' or ``\texttt{Do the people who write this news know the language?}.''

Additionally, $96,717$ reactions ($28.63\%$) are classified as neutral. With reactions like ``\texttt{Motorcycles can be enjoyed in many ways},'' or ``\texttt{Stay calm, until after the February 14 elections and the government is in place, they will not make a single decision}.''

Finally, positive posts are the least common with $29,523$ instances ($8.74\%$). With posts such as ``\texttt{He has made very good programs with intelligent humor},'' or ``\texttt{Therapy is the best thing that has ever happened to me}.'' Therefore, this dataset has a lot of negative content, and only one out of ten reactions has a positive nature.

\begin{table*}[t!]
\centering
\begin{tabular}{c|p{1.5cm}p{1.5cm}p{1.5cm}||p{1.5cm}p{1cm}||p{1.5cm}}
& \multicolumn{3}{c}{\textbf{Sentiments}} & \multicolumn{2}{c}{\textbf{Hate}} & \\
\textbf{Category} & \textbf{Positive} & \textbf{Neutral} & \textbf{Negative} & \textbf{No} & \textbf{Yes} & \textbf{Total}\\
\hline
\hline

\rule{0pt}{2.3ex}\social{Social issues} & $6,648$ (1.97\%) & $21,750$ (6.44\%) & $\textbf{46,650}$ (\textbf{13.81\%}) & $\textbf{71,768}$ (\textbf{21.25\%}) & $3,280$ (0.97\%) & $75,048$ ($22.22\%$)\\
\hline

\rule{0pt}{2.3ex}\expressions{Expressions \& slang} & $6,070$ (1.80\%) & $20,060$ (5.94\%) & $\textbf{42,611}$ (\textbf{12.61\%}) & $\textbf{65,570}$ (\textbf{19.41\%}) & $3,171$ (0.94\%) & $68,741$ ($20.35\%$)\\
\hline

\rule{0pt}{2.3ex}\politic{Political issues} & $3,476$ (1.03\%) & $11,390$ ($3.37$\%) & $\textbf{24,985}$ (\textbf{7.40\%}) & $\textbf{38,328}$ (\textbf{11.35\%}) & $1,523$ (0.45\%) & $39,851$ ($11.80\%$)\\
\hline

\rule{0pt}{2.3ex}\public{Public administration} & $3,469$ (1.03\%) & $11,083$ (3.28\%) & $\textbf{24,953}$ (\textbf{7.39\%}) & $\textbf{38,126}$ (\textbf{11.29\%}) & $1,379$ (0.41\%) & $39,505$ ($11.69\%$)\\
\hline

\rule{0pt}{2.3ex}\spain{Spanish politics} & $3,176$ (0.94\%) & $10,520$ (3.11\%) & $\textbf{23,679}$ (\textbf{7.01\%}) & $\textbf{36,086}$ (\textbf{10.68\%}) & $1,289$ (0.38\%) & $37,375$ ($11.06\%$)\\

\hline
\rule{0pt}{2.3ex}\covid{COVID-19} & $2,204$ (0.65\%) & $7,140$ (2.12\%) & $15,976$ (4.73\%) & $\textbf{24,382}$ (\textbf{7.22\%}) & $938$ (0.28\%) & $25,320$ ($7.50\%$)\\

\hline
\rule{0pt}{2.3ex}\inter{International relations \& immigration} & $2,069$ (0.61\%) & $6,970$ (2.06\%) & $15,449$ (4.57\%) & $23,584$ (6.98\%) & $904$ (0.27\%) & $24,488$ ($7.25\%$)\\

\hline
\rule{0pt}{2.3ex}\media{Hobbies \& lifestyle} & $1,368$ (0.4\%) & $4,491$ (1.33\%) & $9,633$ (2.85\%) & $14,813$ (4.38\%) & $679$ (0.2\%) & $15,492$ (4.58\%)\\

\hline
\rule{0pt}{2.3ex}\transport{Safety \& primary sectors} & $1,043$ (0.31\%) & $3,313$ (0.98\%) & $7,631$ (2.26\%) & $11,538$ (3.41\%) & $449$ (0.13\%) & $11,987$ (3.55\%)\\

\hline
\hline
\textbf{Total} & $29,523$ (8.74\%) & $96,717$ (28.63\%) & \textbf{211,567} \textbf{(62.63\%)} & \textbf{324,195} \textbf{(95.97\%)} & $13,612$ ($4.03\%$) & $337,807$ ($100\%$)\\
\end{tabular}
\caption{Distribution of sentiments and hate by category.}
\label{tab:all-results}
\end{table*}

\subsubsection{Distribution of sentiments in categories}

Table \ref{tab:all-results} presents the distribution of sentiments (positive, negative, and neutral) for each category. Consequently, most topics have a high percentage of negative sentiment compared to positive and neutral sentiment.

\paragraph{Negative public reactions}

Initially, the first two categories are very negative. In particular \social{Social issues} seem to generate the most negative discussions with 46,650 posts (13.81\%). Having topics, such as \social{Gender \& feminism} which is third most toxic, with  $9,897$ (2.84\%) negative messages, like ``\texttt{the pepper spray has run out, almost all the women I know have had to buy it because the nearby police officers all comment the same thing that there is so much going on},'' or ``\texttt{where are the feminists where is the left that claims to defend women?}.'' Another topic in this category is \social{Media} with $5,815$ negative reactions (1.67\%). This topic includes sarcastic or provocative posts like ``\texttt{The manipulation by ABC is textbook},'' which means that the manipulative practices by the media outlet ABC are a perfect example of how manipulation is typically done. With fewer negative reactions, in this category, there is the \social{Cycling \& motorcycling} topic, which is the least with negative reactions $467$ (0.13\%). This topic has posts such as ``\texttt{Cyclists are vandals and thugs because they ride in packs}.''

Following the \social{Social issues} category is the \expressions{Expressions \& slang} category with 42,611 posts (12.61\%). This may reflect the sensitivity and polarization these topics generate in society. In this category, there are topics such as \expressions{Conversation expressions}, which is the most toxic one, with $16,485$ (4.74\%) negative posts such as ``\texttt{what a big lie},'' or ``\texttt{the mother who bore him}.'' Then, the topic \expressions{Negative expressions} is the second most toxic, with $16,034$ (4.61\%). They contain posts with a negative connotation in Spanish, such as ``\texttt{not with my taxes!},'' or ``\texttt{but is everyone an idiot?}.''

There is a group of categories with a similar proportion of negative content (around 7\%). In particular, the \politic{Political issues} category has 24,953 negative posts (7.40\%). Having topics such as \politic{Goverment} with 4,556 negative reactions (1.35\%). Having posts like ``\texttt{this miserable government},'' or ``\texttt{this government will never admit failure or fault, it is always someone else's fault}.'' With fewer negative reactions, there is the \politic{Populism} topic, which is the third topic with less negative reactions, having $492$ negative reactions (0.14\%). This topic includes messages such as ``\texttt{This incompetent populist}.''

Following closely is the \public{Public administration} category with $24,953$ negative reactions (7.39\%). Having topics, like \public{Electoral system} with $4,377$ negative reactions (1.3\%). Including posts such as ``\texttt{the elections should be suspended, there are many people who cannot vote by mail},'' or ``\texttt{to manipulate elections as in the USA a judge should control this}.''

The third group category, comprising around $7\%$ of negative posts, is \spain{Spanish politics}. In particular, it has $23,679$ negative reactions (7.01\%). Including topics, such as \spain{Spanish politics} with $5,531$ (1.59\%) negative messages with posts like ``\texttt{spanish politicians are all at war against Spain they weaken our health our physical capacity our economic capacity our army they destroyed the industry and all sources of financing}.''

Another group with around $4\%$ of negative reactions is formed by two categories. First is the \covid{COVID-19} category, with $15,449$ negative messages (4.73\%). Including topics such as \covid{Vaccination}, with $8,170$ (2.35\%) negative messages. Having negative and sarcastic posts like ``\texttt{who is in charge of controlling that vaccines are not given by friends and family?}.''

The second category of this group is the \inter{International relations \& immigration}, with 15,449 negative reactions (4.57\%). It includes topics such as \inter{Moroccan immigration}, which is the second topic with less negative reactions, having $479$ (0.14\%) negative reactions, like ``\texttt{I don't think they were black; they seemed to be North Africans. They are like a plague}.''

The last group comprises categories with around 2\% of negative content. In particular, the \media{Hobbies \& lifestyle} category has 9,633 negative reactions (2.85\%). It includes topics such as \media{Music} with 2,077 negative reactions (0.62\%) such as ``\texttt{how is he going to know music if he doesn't make music and he is not a musician or anything like that, he is a product that serves to make money from people's money and in a few years nobody will even know who he is.}.''

Lastly, the \transport{Safety \& primary sectors} category has 7,631 negative reactions (2.26\%). It includes topics like \transport{Emergency services} which has 1,208 negative reactions (0.36\%) such as ``\texttt{it is embarrassing to include the miracle in one of the accidents that unfortunately occur}.''

\paragraph{Neutral public reactions}

Neutral content is consistently lower than negative content, often amounting to half of the volume of negative messages.

For instance, the \social{Social issues} category has the most neutral content with $21,750$ neutral reactions (6.44\%). This category includes topics such as \social{Money matters} with $2,086$ neutral messages (0.62\%), such as ``\texttt{the Bank of Spain will allocate almost one million euros to pay for apartments for its employees}.''

Following is the \expressions{Expressions \& slang} category having $20,060$ neutral posts ($5.94\%$) with topics such as \expressions{Sarcastic expressions}, it includes $279$ neutral posts ($0.08\%$) like ``\texttt{Qué huevos tienes}'', which translates to ``\texttt{What eggs you have}'', with derogatory or dismissive connotation.

Moreover, the \politic{Political issues} category has $11,390$ neutral posts (3.37\%), along topics such as \politic{US politics} which has $1,911$ neutral posts (0.57\%) like ``\texttt{Trump will go down in history as the man who could go down in history for fulfilling the legend or myth of pushing the nuclear button}.''

Furthermore, the \public{Public administration} category includes $11,083$ neutral posts (3.28\%) with topics such as \public{Taxation} which has $722$ neutral messages (0.21\%) like ``\texttt{The Treasury has fighter jets and the SEPE has biplanes}.'' It is worth noting that the SEPE, or State Public Employment Service, manages employment services and benefits in Spain.

Finally, the \spain{Spanish politics} category has 10,520 neutral messages (3.11\%) in topics such as \spain{Catalonia} with $2,312$ neutral messages (0.68\%) like ``\texttt{get subtitled in your country XD}'' which is sarcastic about Catalan language.

\paragraph{Positive public reactions}

The percentage of positive content is notably low, typically ranging from one-seventh to one-eighth of the volume of negative content. This disparity highlights a significant tendency toward negative sentiment in the public reactions analyzed.

In particular, the \social{Social issues} category has the most positive content with $6,648$ positive reactions (1.97\%). In topics such as \social{Religion} with $233$ positive reactions (0.07\%) including posts like ``\texttt{thanks be to the Lord. My prayers have been heard}.''

Following, is the \expressions{Expressions \& slang} category having $6,070$ positive posts ($1.80\%$) with topics such as \expressions{Time expressions}, it includes $999$ positive posts ($0.3\%$) like ``\texttt{In these times it is an excellent plan}.''

Moreover, the \politic{Political issues} category has $3,476$ neutral posts (1.03\%), along topics such as \politic{Democracy} with $147$ positive reactions (0.04\%) like ``\texttt{do not confuse the executive power with what each party wants to do in a democracy where there is freedom of thought. Some can attack others with arguments and reason, while others can defend themselves with other arguments and reason}.''

Finally, the \public{Public administration} category includes $3,469$ positive posts (1.03\%) with topics such as \public{Pension system} which has $90$ positive messages (0.03\%) like ``\texttt{Well, I don't know what to tell you, if I prefer to pay taxes today when I am old enough and healthy enough to work more if I need to contribute more than when I am too old}.''

\subsection{\textbf{Objective 3: Hate analysis}}\label{results:hate}

Regarding the prevalence of hate, an analysis is performed on the posts originally labeled with the presence of hate by the Hatemedia project. In this instance, our attention is directed towards the `\textit{Hate}' column in Table \ref{tab:all-results}.

\subsubsection{Distribution of hate in the dataset}

The prevalence of hate speech within public reactions was analyzed, revealing that most posts, totaling $324,195$ instances ($95.97\%$), do not contain hate speech. Conversely, a smaller subset of $13,616$ posts ($4.03\%$) was identified as hateful. This indicates a significant contrast between the high negativity discussed previously and the relatively low presence of hate speech.

In particular, the content labeled as hateful has $12,498$ (3.7\%) reactions with negative sentiment, $338$ (0.1\%) reactions with neutral sentiment, and $776$ (0.23\%) reactions with positive sentiment, indicating that most reactions to hateful content are overwhelmingly negative, with a significant disparity compared to neutral and positive reactions. This consistent sentiment analysis reinforces the observation that the audience predominantly responds negatively to content perceived as hateful.

\subsubsection{Distribution of hate in categories}

The prevalence of hate speech varies across different categories, consistently remaining low. Notably, the occurrence of hate speech partly correlates with the volume of negative messages within each topic. Consequently, the category \social{Social issues} has the most hateful content with $3,280$ instances (0.97\%). Having topics such as \social{Gender \& feminism}, which is the third topic with most hateful content having $705$ ($0.20\%$) messages with hate posts like ``\texttt{she pretends to be different but she is dedicated to show meat like almost all the girls who want to gain followers in the networks the prostitution of the 21st century is online and generalized}.''

The second category with the most hateful content is \expressions{Expressions \& slang} with $3,171$ messages (0.94\%). This category includes the topic with the most hateful content, which is \expressions{Conversation expressions} with $1,284$ hateful posts ($0.37\%$) like ``\texttt{how cheap it is to give a slap}.'' In this category is the second topic with the most hateful content. The \expressions{Negative expressions} topic has $1,107$ ($0.32\%$) hateful reactions such as ``\texttt{you are a very grown-up gentleman, get down from your wooden horse and stop making a fool of yourself}.''

Moreover, the \politic{Political issues} category has $1,523$ hateful messages (0.45\%) along topics such as \politic{Fascism} with $81$ hateful reactions (0.02\%) like ``\texttt{those who are antifascists turn out to be the most fascist and totalitarian headless chickens who know neither what they do nor what they say, they follow the dictates of their leaders without thinking.}.'' On the other hand, in this category, there are topics with a low prevalence of hate. In fact, the \politic{Populism} topic is the third topic with less hateful content having $33$ hateful reactions ($0.01\%$) like ``\texttt{the problem is the populists of the Podemos and the Psoe, not to mention the terrorists and murderers who accompany them in government}.''

In this vein, the \public{Public administration} category has $1,379$ (0.41\%) hateful reactions in topics like \public{Justice system}, which has $364$ messages with hate (0.11\%) such as ``\texttt{Killing and sexually abusing a woman and getting a few years in prison is ridiculous}.'' In contrast, this category includes the second topic with less hateful content. The \public{Transport system} topics with $31$ ($0.009\%$) hateful posts like ``\texttt{this is the only idiot who believed the campaign that the metro flies and thinks the metro is an airport}.''

Furthermore, the \spain{Spanish politics} category has 1,289 hateful reactions (0.38\%) in topics such as \spain{Madrid} with $455$ hateful posts (0.13\%) like ``\texttt{Ayuso resign! Incompetent and puppet of the corrupt PP, She does not represent the people of Madrid},'' referring to the president of Madrid, Isabel D\'iaz Ayuso,

However, some categories with high volumes of negative sentiment do not exhibit a proportional increase in hate speech. For instance, while the category \covid{COVID-19} has $15,976$ negative messages ($4.73\%$), it only shows a hate speech rate of $0.28\%$, indicating that although the discussions are predominantly negative, they do not necessarily translate into hateful content. 

Similarly, \inter{International relations \& immigration} has $15,449$ messages with negative sentiment ($4.57\%$) but a hate speech rate of 0.27\%. These observations suggest that while negative sentiment can contribute to hate speech, other factors, such as the nature of the topic and the context of discussions, significantly influence the extent of hateful content. This category includes topics like \inter{Moroccan immigration}, which is the topic with the least hateful content having $28$ hateful reactions ($0.008\%$) such as ``\texttt{Moroccan of those who come in patera to commit crimes and that you pamphleteers want to hide to continue collecting from the system}.''

\subsection{\textbf{Objective 4: Online reactions to Spanish media characterization}}\label{results:media}

In this step, the labeled data is analyzed by aggregating the messages according to the media outlet to which they react. This approach facilitates the examination of differences in topic distribution, sentiment, and hate speech levels across each Spanish media included in the analysis (\textit{La Vanguardia}, \textit{ABC}, \textit{El Pa\'is}, \textit{El Mundo}, and \textit{20 Minutos}). By analyzing these aggregated responses, potential correlations and dependencies between the news published by these outlets and the public's reactions on social media can be identified.

The dataset's content distribution across different media outlets reveals the number of reactions derived from each media outlet. However, this does not necessarily indicate the volume of content each medium generates due to potential collection biases. Notably, \textit{El Mundo} has the most reactions with $139,708$ messages (41.36\%), followed by \textit{ABC} with $69,878$ messages (20.69\%), \textit{20 Minutos} with $48,315$ messages (14.30\%), \textit{El Pa\'is} with $38,102$ messages (12.28\%), and \textit{La Vanguardia} with $39,610$ messages (11.37\%). 

\subsubsection{Distribution of topics in Spanish media}

First, the approach taken to understand the variations in topic distribution following the methodology stated in Section \ref{met:topic-media}, due to the complexity of comparing $81$ topics across each media outlet, is as follows. The first step involves calculating the distribution of standard deviations for each topic within a media outlet compared to the others. Specifically, for each media outlet, and for each topic, the standard deviation between the number of messages generated by that media outlet and the average number of messages from the other outlets is calculated. 

\begin{figure}[h!]
	\centering{
		\includegraphics[width=0.8\columnwidth]{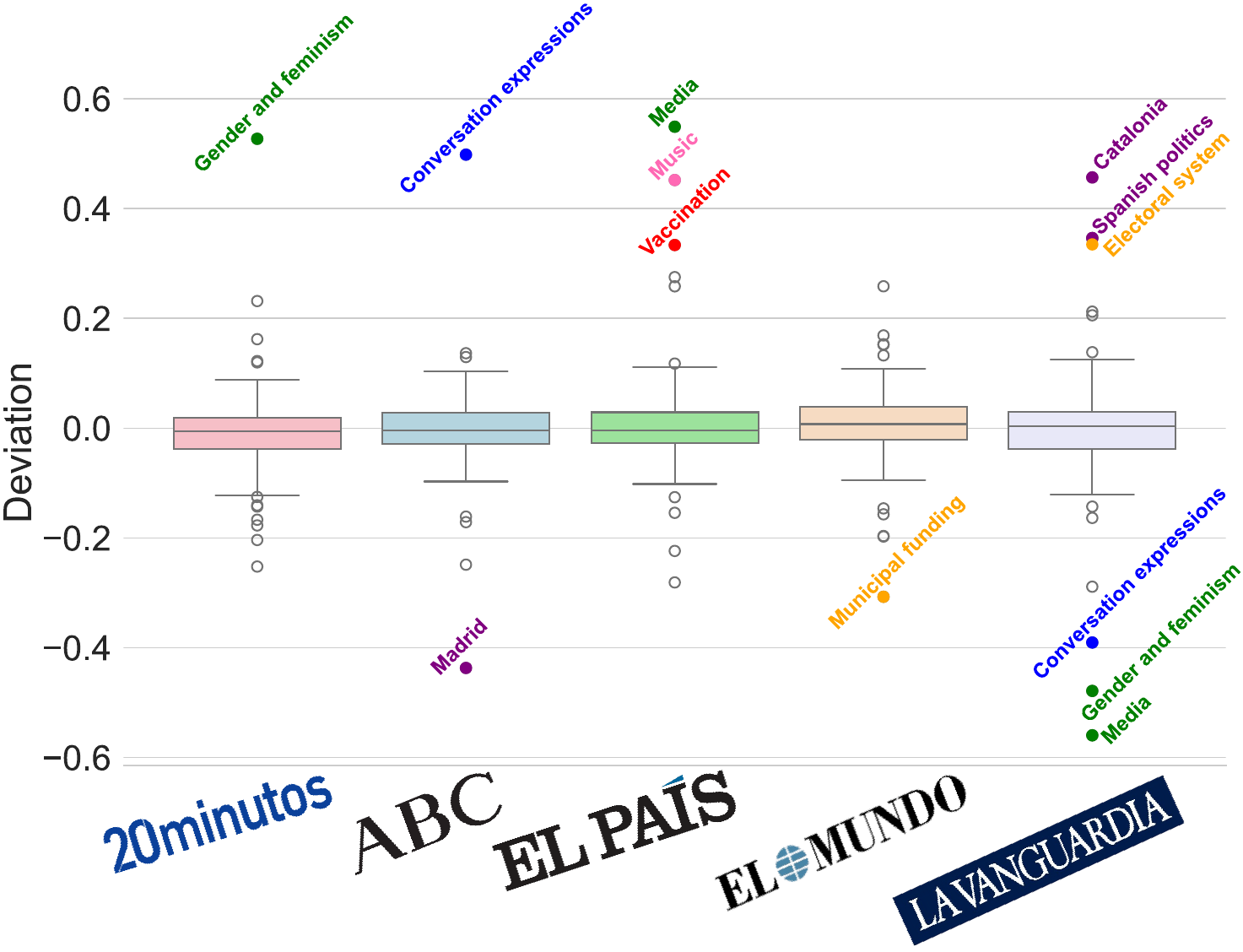}
	\caption{Distribution of absolute deviation of topics per media outlet concerning the rest.}\label{fig:media-topics-box}
 }
\end{figure}

In this context, Figure \ref{fig:media-topics-box} illustrates the variability in the distribution of posts across the $81$ topics for each media outlet. Accordingly, most topic categories exhibit a low deviation, indicating that the values are closely clustered around the median. Therefore, the variations within each media outlet are minimal, highlighting a consistent pattern in how topics are covered.

On the other hand, the outliers with more than $0.3$ of deviation or less than $-0.3$ in the graph are clearly marked and labeled with different colors according to their respective categories. These points represent categories where a media outlet generates significant reactions concerning the rest of the firms. This discrepancy could be attributed to the nature and framing of the news and the volume of news articles published within each topic, influencing the overall reaction count. 

In particular, the topic \social{Gender \& feminism} shows a deviation greater than $0.5$ in \textit{20 Minutos}, indicating that this media outlet generates a broader range of reactions to this particular topic than the other firms. Notably, \textit{20 Minutos} does not have a clearly marked political orientation, which might contribute to the diverse range of reactions it generates on such socially sensitive topics.

Moreover, in \textit{ABC}, the topic \expressions{Conversation expressions} exhibit a deviation greater than $0.4$. While the topic \spain{Madrid} has a deviation lower than $-0.3$, indicating a less polarized response from the audience in this topic. Therefore, this suggests that \textit{ABC} may be raising more reactions around informal expressions, while its coverage of \spain{Madrid} appears to generate fewer reactions, potentially aligning with the outlet's conservative ideology.

In \textit{El País}, which has shown an editorial line close to the Spanish political party PSOE with social-democratic tendencies, the \covid{Vaccination} topic is an outlier, exhibiting a deviation greater than $0.3$. Similarly, the \media{Music} topic shows a significant deviation exceeding $0.4$. The \social{Media} topic also stands out with a deviation greater than $0.5$. Furthermore, in \textit{El Mundo}, only the topic \public{Municipal funding} is an outlier with a deviation slightly lower than $-0.3$.

Lastly, in \textit{La Vanguardia}, the distribution of topics reveals a distinct focus on several political and social areas, which are marked by significant deviations, indicating diverse and polarized reactions from the audience. The \social{Media} topic stands out with a deviation lower than $-0.5$. Additionally, the topic \spain{Catalonia} shows deviations greater than $0.4$, reflecting the intense interest and likely divisive opinions surrounding Catalan independence and related political issues, which are particularly relevant to \textit{La Vanguardia}’s readership, since this media has Catalanism as their ideology. While the topic \social{Gender \& feminism} has a deviation greater than $0.5$. Finally, the topic \expressions{Conversation expressions} has a deviation lower than $-0.3$. The topic \spain{Spanish politics} has a deviation greater than $0.3$, highlighting that political discourse is a significant driver of engagement among \textit{La Vanguardia}’s audience, suggesting that the outlet generates more reactions in discussions about politics.

Therefore, most topics across different media outlets have low deviations, indicating a consistent pattern in how topics generated reactions. However, significant outliers in specific categories highlight areas where media outlets diverge. Notably, topics like gender and feminism, media, and local political issues such as Catalonia and Madrid raise varied responses, underscoring their complexity and the broad spectrum of public opinion associated with them.

\subsubsection{Distribution of sentiments in Spanish media}

Table~\ref{tab:media-sent-hate} presents the distribution of sentiments (positive, neutral, and negative) and the presence of hate content (no, yes) across Spanish media outlets. The table provides both the absolute counts and relative percentages within each sentiment and hate category, as well as the total counts and absolute percentages for each media outlet.

The sentiment analysis reveals a predominant negative sentiment across all media outlets. \textit{El Mundo} shows the highest proportion of negative sentiment (28.2\%), followed by \textit{ABC} (13.14\%), \textit{20 Minutos} (7.59\%), \textit{El Pa\'is} (7.58\%),  and \textit{La Vanguardia} (6.12\%). On the other hand, \textit{El Mundo} has the highest percentage of positive sentiment (3.26\%), neutral content (9.89\%)

The high prevalence of negative sentiment in public reactions suggests that the news content covered by the media often elicits critical or adverse responses from the audience. This indicates that the public may perceive the topics discussed in the news as more pessimistic or concerning, which can shape the general mood and discourse on social media and web platforms. It is important to note that this observation reflects the reactions to the media content rather than the tone or nature of the media coverage, which was not directly analyzed in this study.

\subsubsection{Distribution of hate in Spanish media}

Regarding hate speech, Table~\ref{tab:media-sent-hate} presents the distribution of hate content and reveals that reactions derived from Spanish media content are not classified as hate, with percentages ranging from $93.97\%$ to $98.7\%$ for non-hate content.

Regarding the hateful reactions, \textit{El Mundo} raises the highest percentage of hate content ($1.65\%$), followed by \textit{20 Minutos} ($0.86\%$) and \textit{ABC} ($0.71\%$). On the other hand, \textit{La Vanguardia} triggers the lowest percentage of hate content ($0.15\%$). This data suggests that explicit hate content is comparatively rare, while negative sentiments are prevalent. This may be a consequence of the polarization and tension within the audience, which is often reflected in the media publications themselves. These outlets typically cover political news and current events that can cause strong reactions, as observed in the public reactions analyzed in our study.

\begin{table*}[ht!]
\centering
\small
\begin{tabular}{p{2.05cm}|ccc||cc||p{1cm}}
& \multicolumn{3}{c}{\textbf{Sentiments}} & \multicolumn{2}{c}{\textbf{Hate}} & \\
\textbf{Media} & \textbf{Positive} & \textbf{Neutral} & \textbf{Negative} & \textbf{No} & \textbf{Yes} & \textbf{Total}\\
\hline
\hline

\textit{El Mundo} & $10,997$ (3.26\%) & $33,460$ (9.89\%) & $95,251$ (\textbf{28.2\%}) & $134,120$ (39.71\%) & $5,588$ (1.65\%) & \textbf{139,708} ($41.36\%$)\\
\hline

\textit{ABC} & $5,693$ (1.69\%) & $19,785$ (5.86\%) & $44,400$ (13.14\%) & $67,503$ (19.98\%) & $2,375$ (0.71\%) & \textbf{69,878} ($20.69\%$)\\
\hline

\textit{20 Minutos} & $4,339$ (1.28\%) & $18,340$ (\textbf{5.44\%}) & $25,636$ (7.59\%) & $45,400$ (13.44\%) & $2,915$ (\textbf{0.86\%}) & \textbf{48,315} ($14.30\%$)\\
\hline

\textit{El Pa\'is} & $4,197$ (1.24\%) & $12,019$ (3.56\%) & $25,588$ (7.58\%) & $39,570$ (11.71\%) & $2,234$ (0.66\%) & \textbf{38,102} ($12.28\%$)\\
\hline

\textit{La Vanguardia} & $4,297$ (\textbf{1.27\%}) & $13,113$ (3.88\%) & $20,692$ (\textbf{6.12\%}) & $37,602$ (11.13\%) & $500$ (0.15\%) & \textbf{39,610} ($11.37\%$)\\

\hline
\hline
\textbf{Total} & $29,523$ (8.74\%) & $96,717$ (28.63\%) & 211,567 (62.63\%) & 324,195 (95.97\%) & $13,612$ ($4.03\%$) & $337,807$ ($100\%$)\\
\end{tabular}
\caption{Distribution of sentiments and hate in reactions caused by media publications.}
\label{tab:media-sent-hate}
\end{table*}

\section{Conclusion and future work}\label{conclusion}

This study is particularly centered on the landscape of Spanish media. It involves the analysis of $337,807$ posts comprising public reactions derived from the Spanish media outlets' content and sourced from the Hatemedia project. These reactions were gathered from $\mathbb{X}$ (formerly Twitter) and the websites of five prominent Spanish media outlets, namely, \textit{La Vanguardia}, \textit{ABC}, \textit{El Pa\'is}, \textit{El Mundo} and \textit{20 Minutos}.

These data are used for topic modeling, specifically utilizing BERTopic. Consequently, the BERTopic model was fine-tuned, identifying 81 distinct topics, each representing specific themes within the dataset, such as conversation expressions, gender and feminism, media, vaccination, or government. The labeling process involved four LLMs to assist the authors in the topic labeling. In this context, it was observed that Llama2 provided more abstract labels. The topics were classified into nine categories: Social issues, expressions and slang, political issues, public administration, Spanish politics, COVID-19, international relations and immigration, hobbies and lifestyle, and safety and primary sector. We identify that the category of social issues has the most associated content, being the most popular, and encompasses the most topics, being the most variate with the highest number of subtopics associated, followed by expressions and slang, political issues, public administration, and Spanish politics. This indicates the complexity and scope of these particular topics and the tendency to discuss social issues in social media.

Following the application of BERTopic, sentiment analysis was conducted to categorize the posts into positive, negative, or neutral sentiments. To achieve this, the \textit{TweetNLP} tool was utilized. The findings indicate a significant prevalence of negative content within the topics, whereas positive content was notably less frequent. Additionally, there was a noticeable variation in negative content across different topics. For example, topics related to expressions, politics, gender issues, and COVID-19 had the most negative content. The negative nature of these themes could relate to the radicalization, polarization, and the well-known \textit{infodemic} of conspiracy theories and disinformation phenomena. 

Additionally, the study examined the presence of hate speech within the topics. Analysis revealed that a minor portion of posts included hateful content ($4.1\%$), with the topics showing the highest levels of negativity also exhibiting the greatest concentration of hate speech. This subset of the dataset is characterized by aggressive and negative language, which may have a substantial emotional influence on the digital media landscape. Notably, sensible topics in the Spanish context related to government critiques and public perception of politicians were found, potentially contributing to polarization and lack of constructive dialogue, creating a less tolerant and more polarized societal environment.

Finally, a comparison between content from web portals and Twitter was tested, revealing no significant differences in the distribution of topics, sentiments, or hate content. Subsequently, the analysis focused on individual media outlets. When examining the distribution of topics, a consistent pattern in how topics generate reactions is observed. Concerning sentiments, results demonstrate a predominant negative sentiment across all outlets, with \textit{El Mundo} showing the highest proportion of negative content. In contrast, \textit{La Vanguardia} has a more balanced distribution of positive and neutral sentiments, and \textit{20 Minutos} features the most neutral content. Lastly, regarding hate speech, the majority of content across all media is not classified as hate. However, \textit{20 Minutos} contains the highest percentage of hate content, followed by \textit{El Mundo} and \textit{ABC}, while \textit{La Vanguardia} has the least. These findings indicate that hate content is less frequent while negative sentiments are common.

In conclusion, the analysis reveals that no specific media outlet significantly influences audience reactions. Instead, the data demonstrate that the audience itself tends to react negatively across various topics and media. This general negativity likely reflects the intense political and social climate during the period studied, characterized by the COVID-19 pandemic, the Catalonia independence movements, and significant climatological events in Spain. Media coverage mirrors the social and political reality, providing insights into the audience's sentiments within the country's socio-political context. Therefore, the minimal differences between media outlets support the hypothesis that irrespective of the medium, the audience exhibits consistent negativity and toxicity across all platforms and topics.

Future work could focus on extending the study period to capture long-term trends and incorporating a wider range of media sources, including regional and alternative outlets. Additionally, a multilingual approach encompassing Spain's diverse linguistic landscape could provide deeper insights into regional differences in public sentiment. Investigating the impact of disinformation on public opinion and comparing these findings with international data could offer valuable perspectives and best practices for mitigating online toxicity and enhancing the quality of public discourse. These insights can inform policy recommendations and industry practices for responsible media engagement and content moderation.

\section*{Acknowledgment}
This work has been partially funded by the strategic project CDL-TALENTUM from the Spanish National Institute of Cybersecurity (INCIBE) and the Recovery, Transformation, and Resilience Plan, Next Generation EU, and the University of Murcia with the FPI/0000902983 contract.

We thank the Hatemedia project (PID2020-114584GB-I00), financed by MCIN/AEI/10.13039/501100011033, for providing the dataset used in this work.

\bibliographystyle{model2-names}\biboptions{authoryear}
\bibliography{bibliography.bib}
\end{document}